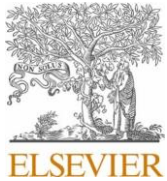
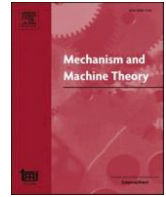

Research paper

# An Analysis of Higher-Order Kinematics Formalisms for an Innovative Surgical Parallel Robot

Calin Vaida[a], Iosif Birlescu[a], Bogdan Gherman[a], Daniel Condurache[a,b,c], Damien Chablat[a,d], Doina Pisla[a,c*]

[a] *Research Center for Industrial Robots Simulation and Testing, Technical University of Cluj-Napoca, Memorandumului 14, 400114 Cluj-Napoca, Romania*

[b] *Gheorghe Asachi Technical University of Iasi, D. Mangeron 59, 700050 Iasi, Romania*

[c] *Technical Sciences Academy of Romania, B-dul Dacia, 26, 030167 Bucharest, Romania*

[d] *École Centrale Nantes, Nantes Université, CNRS, LS2N, UMR 6004, F-44000 Nantes, France*

[*] *Corresponding author: doina.pisla@mep.utcluj.ro*

ABSTRACT

The paper presents a novel modular hybrid parallel robot for pancreatic surgery and its higher-order kinematics derived based on various formalisms. The classical vector, homogeneous transformation matrices and dual quaternion approaches are studied for the kinematic functions using both classical differentiation and multidual algebra. The algorithms for inverse kinematics for all three studied formalisms are presented for both differentiation and multidual algebra approaches. Furthermore, these algorithms are compared based on numerical stability, execution times and number and type of mathematical functions and operators contained in each algorithm. A statistical analysis shows that there is significant improvement in execution time for the algorithms implemented using multidual algebra, while the numerical stability is appropriate for all algorithms derived based on differentiation and multidual algebra. While the implementation of the kinematic algorithms using multidual algebra shows positive results when benchmarked on a standard PC, further work is required to evaluate the multidual algorithms on hardware/software used for the modular parallel robot command and control.

## 1. Introduction

Pancreatic cancer is one of the deadliest forms of cancer with a 5-year survival rate of less than 10%, making it the 10th leading cause of death worldwide [1]. The scientific literature reports quite different survival rates in single-center studies which are influenced by many uncontrollable factors, but all agree that aggressiveness, diagnosis in late stages and the complexity of the surgical procedure are limiting the prognosis of this disease. As pancreatic cancer evolves insidiously, at the time of diagnosis less than 25% of the tumors meet resectability criteria, and even these patients have poor prognosis due to the complexity of the surgical treatment – *Whipple procedure* - which implies extensive dissections and multiple anastomosis, with risks of further postoperative complications [2]. As surgery remains the only viable treatment option, it is only natural to make efforts in maximizing its outcomes, whereas the use of robotic systems can play a decisive role in this task. From targeted radiotherapy, robotic-assisted intraoperative ultrasound with CT/MRI data fusion and assistive tasks in the most difficult procedural steps (which often lead to post-operative complications) are just some of the areas of intervention towards a better prognosis of pancreatic cancer [3].

An important aspect of surgical robots which manipulate instruments in minimally invasive procedures refers to the behavior of the instrument with respect to the insertion point in the patient body, which is named the Remote Center of Motion (RCM) point [4]. Kinematically, this point is a class 2 joint (restricting two motions) which can be compensated in several ways: using specific architectures (i.e. spherical or parallelogram), mechanically constraining this point (adding an active 2 DoF (degrees of freedom) joint) or, for simple non-contact tasks a 2-DoF passive joint [5]. Various robotic mechanisms with Remote Center of Motions (RCM) architectures have been analyzed in [5-11].

While the scientific community has acknowledged the importance of the da Vinci surgical platform in robotic surgery, many new parallel or hybrid mechanisms have been proposed as alternatives for a specific procedure due to their advantages in high positioning accuracy and a limited (smaller) workspace which becomes an important safety feature in specific areas of the body [12,13].

A novel hybrid parallel robot was recently introduced [14] which is designed for minimally invasive pancreatic surgery, where the task of the robot is to replace a second surgeon and perform different assistive tasks during the surgery, such as tissue manipulation during tissue and vascular dissection, intraoperative ultrasound for accurate tumor localization. This in turn, allows for adequate space and facilitates vision for the surgeon which operates manually.

Higher-order kinematics is required for such a robot to manipulate the tissue using smooth trajectories. The standard actuators (e.g., servo rotary motors) nowadays are jerk limited, meaning that they can be controlled using trapezoidal acceleration profiles.

As a general overview higher-order kinematics have been studied to improve trajectory planning and accuracy (smoothing the abrupt changes in acceleration, jerk and so on), motion control for adaptive and dynamic environments, stability and energy efficiency, vibration control and autonomous systems. The development of high-precision mechanical systems, of space procedures for docking [15,16] advanced control of robots for precise trajectory tracking [17-20], machine vision [21, 22], biomechanics [23, 24] requires the computation of higher-order kinematics, especially acceleration, jerk or jounce. Where higher-order motion continuity and improved control in robotic applications is required, feedback from higher-order joints accelerations is also necessary. Consequently, a kinematic mapping is required to relate the motion of a robot's end-effector, or more generally, the motion of a rigid body to the motions of a kinematic chain. Higher-order kinematics are developed through computing the time-derivatives of the twist components of the kinematic chain. Each formalism for higher-order kinematics has specific advantages and applications. **Vector-based methods** [25,26] are intuitive and commonly used for simple systems or linear motions but are less effective in handling rotations or constraints; the higher-order kinematics is derived through differentiation of the matrix expressions. **Screw theory methods** [27] are very efficient in scenarios where rotational and translational motions are tightly coupled, such as robotic manipulators, and are effective for singularity analysis and kinematic optimization. Higher-order kinematics for lower-pair joints has been investigated in [28], stressing that the derivative of an arbitrary order can be represented in a compact way using the Lie brackets, greatly increasing the compactness of the Jacobian through expressing the partial derivatives of the instantaneous joint screws. **Dual quaternion methods** [29], integrate rotations and translations into a compact representation, are widely applied in areas like computer graphics, virtual reality, and robotics with complex motion constraints, offering efficient trajectory optimization. The compact representation of dual quaternion parametrization of SE(3) reduces computational complexity while maintaining numerical stability, making it particularly effective for applications that require efficient and accurate computation of rigid body motions [30,31]. **Dual number methods** [32] are known for computational simplicity and useful for deriving simultaneously the higher-order kinematics of lower-pair chains. A first comparative performance analysis over dual orthogonal matrices, dual unit quaternions, dual special unitary matrices and dual Pauli spin matrices was performed in [33]. Moreover, previous work [32,34-36] shows that multidual algebra represents a unifying framework for computing the higher-order kinematics of lower-pair chains, i.e., the multidual operator can be applied in various formalisms, for automatic differentiation, yielding simultaneously the higher-order kinematics up to a degree n.

While the mathematical models used in the kinematic functions for the control system cannot influence directly the outcomes of pancreatic surgery, lower computation times allow the introduction of supplementary elements, (force/torque control, data fusion) performed in real-time, which, on the overall can enhance the critical steps of the surgery leading towards fewer complications and thus, potentially better short- and long-term outcomes.

The aim of this work is to use multidual algebra to derive algorithms for higher-order kinematics for the modular surgical robot introduced in [14], and to compare them with algorithms obtained through classical differentiation methods. Three different formalisms are studied based on: i) geometric/vector method, ii) homogeneous transformation matrices method, iii) dual quaternion method. The motivation is twofold: a) shown how the multidual algebra is used to compute symbolically the higher-order kinematics for the modular surgical robot; b) study the efficiency of the developed algorithms from a numeric point of view, where the algorithms are compared based on their execution times, and result accuracy. The contributions with respect to the state of the art are the following:

1) The paper shows how multidual algebra is used as a unifying framework for the development of higher-order kinematics on a novel hybrid parallel robot for pancreatic surgery. Three formalisms are studied in conjuncture with different techniques derived from multidual algebra.
2) The paper presents initial numerical results regarding the efficiency of the kinematic algorithms, computed on a standard PC (12th Gen Intel(R) Core(TM) i9-12900K  3.20 GHz, 32.0 GB RAM, running Windows 10 Pro).

Section 2 presents the mathematical background used to solve the higher-order kinematics of the proposed parallel robot architecture. Section 3 presents a novel parallel robot designed for pancreatic surgery, with the geometric models derived with the classical vector approach, the dual quaternion approach and the multidual algebra approach. Section 4 presents the higher-order forward and inverse kinematics algorithms. Section 5 provides a statistical analysis for the derived kinematic algorithms and presents the results of a kinematic simulation for a pre-defined trajectory. Section 6 presents a discussion about the results and lastly, the conclusions are drawn in Section 7.

## 2. Mathematical background

Multidual algebra is a unifying framework to compute simultaneously all the higher-order kinematics (up to a degree n) of a rigid body, and it is used in this paper to study the kinematics of the proposed MIS hybrid parallel robot. Multidual algebra can be used with various formalisms (vector methods based on the kinematic Jacobians, homogeneous transformation matrix methods, and dual quaternion methods). The key ingredient is the multidual differential transform (see Appendix A) operator applied to a quantity $S$ " $\overset{(}{S}$ ":

$$\overset{(}{S} = S + \varepsilon \dot{S} + \frac{\varepsilon^2}{2!}\ddot{S} + K + \frac{\varepsilon^n}{n!}S^{(n)} = e^{\varepsilon D}S, \ n \in \mathbb{N}^*, \quad (1)$$

where $S$ is n-times differentiable with respect to time, and $D = \frac{d()}{dt}$ is the time differential operator. It is assumed to be one of the following: a function $f(t)$, a vector $\mathbf{V}(t)$, a tensor $\mathbf{M}(t)$, or a unit dual quaternion $\underline{\mathbf{Q}}(t)$. More details regarding the multidual algebra operators, multidual functions, vectors and tensors are found in Appendix A. This section highlights only the essential results which are used in this paper for deriving higher-order kinematics of robotic systems.

*2.1. Higher-order kinematics using the vector method and multidual algebra*

In the classical vector method [25], the input-output equations describe the mathematical link between the input parameters (i.e. the active generalized coordinates of the robot) and the output parameters (usually parameters describing SE(3) motion). The kinematic equation for velocities is represented by:

$$\mathbf{A} \cdot \dot{\mathbf{X}} + \mathbf{B} \cdot \dot{\mathbf{Q}} = 0, \quad (2)$$

where $\mathbf{A}$ and $\mathbf{B}$ are referred to as the serial and parallel Jacobian matrices [26], which can be derived from the input-output equations of a mechanism, $\dot{\mathbf{Q}}$ is the vector of the active joints velocities (input) and $\dot{\mathbf{X}}$ is the vector of the output generalized velocities. Assuming that $\mathbf{A}$ and $\mathbf{B}$ are nonsingular matrices, the Jacobian $\mathbf{J}$ can be defined as $\mathbf{J} = -\mathbf{B}^{-1} \cdot \mathbf{A}$; the active joint velocities and the output generalized velocities are computed using:

$$\dot{\mathbf{Q}} = \mathbf{J} \cdot \dot{\mathbf{X}},$$
$$\dot{\mathbf{X}} = \mathbf{J}^{-1} \cdot \dot{\mathbf{Q}}. \quad (3)$$

Using this formalism, the higher-order kinematics is obtained through the differentiation (with respect to time) of Eq. (2), leading to mathematical expressions for jerk, jounce or other higher-order kinematic field. The general forms for the n[th] time derivatives (assuming they exist), for the inverse and forward kinematics, are:

$$\mathbf{Q}^{(n+1)} = \sum_{k=0}^{n} \binom{n}{k} \mathbf{J}^{(k)} \cdot \mathbf{X}^{(n-k+1)},$$
$$\mathbf{X}^{(n+1)} = \sum_{k=0}^{n} \binom{n}{k} (\mathbf{J}^{-1})^{(k)} \cdot \mathbf{Q}^{(n-k+1)}, \qquad (4)$$
$$n \in \mathbb{N}^{*}$$

This formalism requires a sequential computation of all the time derivatives up to the $n^{th}$ one.

An alternative method to the classical vector one, for computing higher-order kinematics is to use multidual algebra for automatic differentiation [36]. The multidual operator (Eq. 1) is applied on the kinematic Jacobian $\mathbf{J}$ and the velocity vectors $\dot{\mathbf{x}}$ and $\dot{\mathbf{Q}}$ yielding:

$$\breve{\mathbf{J}} = \mathbf{J} + \varepsilon \dot{\mathbf{J}} + \frac{\varepsilon^2}{2!}\ddot{\mathbf{J}} + \mathbf{K} + \frac{\varepsilon^n}{n!}\mathbf{J}^{(n)},$$
$$\breve{\dot{\mathbf{Q}}} = \dot{\mathbf{Q}} + \varepsilon \ddot{\mathbf{Q}} + \frac{\varepsilon^2}{2!}\dddot{\mathbf{Q}} + \mathbf{K} + \frac{\varepsilon^n}{n!}\mathbf{Q}^{(n+1)}, \; n \in \mathbb{N}^{*}. \qquad (5)$$
$$\breve{\dot{\mathbf{X}}} = \dot{\mathbf{X}} + \varepsilon \ddot{\mathbf{X}} + \frac{\varepsilon^2}{2!}\dddot{\mathbf{X}} + \mathbf{K} + \frac{\varepsilon^n}{n!}\mathbf{X}^{(n+1)},$$

Computing the higher-order invers kinematics is achieved through:

$$\breve{\dot{\mathbf{Q}}} = \breve{\mathbf{J}} \cdot \breve{\dot{\mathbf{X}}}, \qquad (6)$$

considering the required multiplication rules $\varepsilon^k \neq 0, k = \overline{1,n}, \varepsilon^{n+1} = 0, n \in \mathbb{N}$ ; Eq. (6) achieves the (simultaneous) computation of higher-order inverse kinematics of a robotic system.

### 2.2. *Higher-order kinematics using multidual algebra and homogeneous transformation matrices*

In the case of tensors, one can consider a homogenous matrix which describes a rigid motion on a curve in the Lie group of rigid displacements SE(3), [35]:

$$\mathbf{b} = \begin{bmatrix} \mathbf{R} & \mathbf{r} \\ \mathbf{0} & 1 \end{bmatrix}, \mathbf{R} \in SO(3), \mathbf{R} = \mathbf{R}(t), \mathbf{r} = \mathbf{r}(t). \qquad (7)$$

The vector field of higher-order kinematics is:

$$\mathbf{H}_n = \begin{bmatrix} \mathbf{K}_n & \mathbf{v}_n \\ \mathbf{0} & 1 \end{bmatrix}, \qquad (8)$$

where the tensor $\mathbf{K}_n$ and the vector $\mathbf{v}_n$ are obtained from:

$$\mathbf{K}_n = \mathbf{R}^{(n)} \mathbf{R}^T, \; \mathbf{v}_n = \mathbf{r}^{(n)} - \mathbf{K}_n \mathbf{r}. \qquad (9)$$

where $\mathbf{a}_n(\mathbf{p}) = \mathbf{v}_n + \mathbf{K}_n \mathbf{p}$ is the higher-order acceleration vector field, $\mathbf{p} \in \mathbb{R}^3$. Considering the multidual algebra approach, the tensor of higher-order kinematics is obtained from:

$$\breve{\mathbf{H}} = \begin{bmatrix} \breve{\mathbf{K}} & \breve{\mathbf{v}} \\ \mathbf{0} & 1 \end{bmatrix}, \qquad (10)$$

where:

$$\breve{\mathbf{K}} = \breve{\mathbf{R}} \mathbf{R}^T, \; \breve{\mathbf{v}} = \breve{\mathbf{r}} - \breve{\mathbf{R}} \mathbf{R}^T \mathbf{r}, \qquad (11)$$

where $\breve{\mathbf{a}}(\mathbf{p}) = \breve{\mathbf{v}} + \breve{\mathbf{K}} \mathbf{p}$ is the compact form of the higher-order acceleration field expressed using multidual algebra, $\mathbf{p} \in \mathbb{R}^3$. Eqs. (10) and (11) achieve, again, the simultaneous computation of the higher-order kinematics using the homogeneous transformation matrix formalism.

### 2.3. Higher-order kinematics using hyper multidual quaternions

As shown in [37], any Euclidean displacement $D \in SE(3)$ can be mapped into a point $P$ from the projective space $P^7$ through the mapping $m$:

$$m : D \to P \in P^7$$
$$\underline{\mathbf{Q}}(x_i, y_i) \to [x_0 : x_1 : x_2 : x_3 : y_0 : y_1 : y_2 : y_3]^T \neq [0:0:0:0:0:0:0:0]^T \quad (12)$$

with at least one $x_i \neq 0$ required to avoid the trivial solution (which has no use in kinematics). The $x_i : y_i$ parameters are also known as the Study parameters which are based on a unit dual quaternion; the $x_i$ parameters encode rotational displacement whereas $y_i$ parameters encode translational displacement, and they have the structure:

$$\underline{\mathbf{Q}} = \left(1 + \varepsilon_0 \tfrac{1}{2}\mathbf{r}\right)\mathbf{q},$$
$$\mathbf{q} = x_0 + ix_1 + jx_2 + kx_3,$$
$$\mathbf{r} = iy_1 + jy_2 + ky_3, \quad (13)$$
$$\varepsilon_0 \neq 0, \varepsilon_0^2 = 0.$$

which is homomorphic with SE(3) as the unit dual quaternion is a two wrap of the SE(3).

To compute the n-th time derivative of a dual quaternion $\underline{\mathbf{Q}}$ the following relation is used:

$$\underline{\dot{\mathbf{Q}}} = \dot{\mathbf{q}} + \varepsilon_0 \tfrac{1}{2}\left(\dot{\mathbf{r}}\mathbf{q} + \mathbf{r}\dot{\mathbf{q}}\right),$$
$$\underline{\mathbf{Q}}^{(n)} = \frac{d^{n-1}}{dt^{n-1}}\underline{\dot{\mathbf{Q}}}, \quad (14)$$
$$n \in \mathbb{N}^*, \varepsilon_0 \neq 0, \varepsilon_0^2 = 0.$$

If $\underline{\overset{(}{u}}$ is the set of unit multidual quaternions and $\underline{\overset{(}{u}}$ is the set of units HMD quaternions, the following is true [34,35]:

$$\underline{\overset{(}{\mathbf{Q}}} = \left(1 + \varepsilon_0 \tfrac{1}{2}\overset{(}{\mathbf{r}}\right)\overset{(}{\mathbf{q}},$$
$$\varepsilon_0 \neq 0, \varepsilon_0^2 = 0. \quad (15)$$

### 2.4. Discussion on multidual algebra approach for higher-order kinematics

Multidual algebra can be used as a unifying frame to compute higher-order kinematics of robotic systems, simultaneously for all higher kinematics fields. The multidualization is valid for any of the discussed formalisms, the difference being the symbolic quantity that is multidualized.

Using multidual algebra does not affect the known advantages and drawbacks of each of the discussed formalisms:

- The vector approach based on the Jacobian matrices has the advantage of also yielding the input and output singularities trough $\det(\mathbf{A}) = 0, \det(\mathbf{B}) = 0$.
- The homogeneous matrix approach has the advantage of easy programing implementation.
- The quaternion approach is computationally efficient and free of gimbal lock singularities.

However, multidual algebra approaches show two advantages:
1. From a symbolic calculus point of view, multidual algebra computes the higher-order kinematics simultaneously.
2. From a computation point of view, multidual algebra can be used to implement kinematic functions using a programming approach based on simple commutations of the values of $\varepsilon^n / n!$ between the values 0 and 1, which in turn yields the higher-order kinematics; this requires function handlers (e.g., in MATLAB) or function pointers (e.g., in C). Section 4 shows how such kinematic functions are designed and their efficiency over the classical approaches.

## 3. An innovative hybrid parallel robot for minimally invasive surgery

The paper introduces a modular surgical robot, [14], designed for minimally invasive pancreatic surgery, for which the higher-order kinematics are derived further. The surgical robot task is to replace a second surgeon in the operating room, which manipulates specific tissue in the operating field using a surgical tool; this task in turn creates the appropriate workspace for the main surgeon which performs manually the surgical intervention. Higher-order kinematics plays an important role in robotic assisted surgical procedures, especially for defining smooth trajectories, which improve patient safety. Figure 1 presents the CAD model of the 4 DOF modular parallel robot in the medical environment, prepared for the pancreatic minimally invasive surgery.

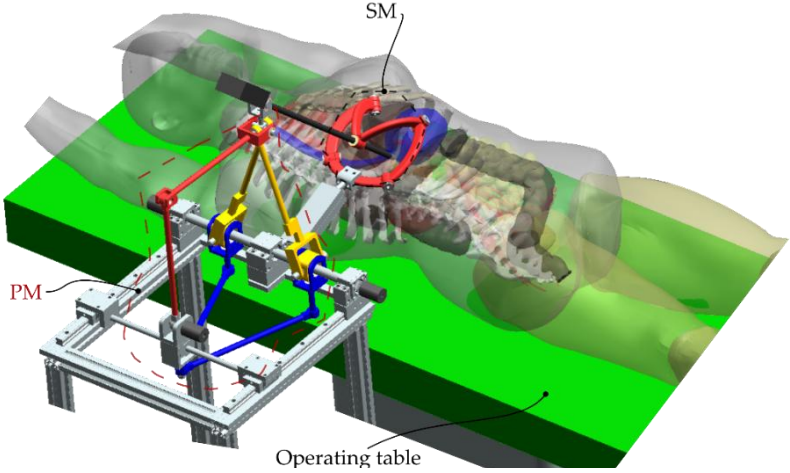

**Figure 1**. The CAD model of the 4-DOF surgical parallel robot in medical environment.

The novel hybrid parallel robot features modular architecture, consisting of two parallel modules:
1) A passive spherical five-bar mechanism (SM) with architecturally constrained Remote Centre of Motion (RCM). One passive revolute joint is replaced with a passive cylindrical joint to allow a surgical instrument (SI) free translation. Consequently, SM allows 4 DOFs 3R+1T, i.e., three orientations of the surgical instrument relative to a point (the insertion point) and one translation along the instrument rod (for the surgical tool insertion and retraction).
2) An active 4 DOFs parallel robot (PM) for the manipulation of SI (which is oriented using SM).

The targeted advantages of the proposed modular robotic architecture versus an active parallel spherical mechanism (e.g. actuated SM) include:

1. The possibility of using long surgical instruments for MIS, which otherwise would be difficult to handle using an active SM alone due to high inertias;
2. The possibility of fast switching from the robotic control to fully manual control of the medical instrument, by removing the PM;
3. An appropriate size of the SM (as small as possible to fit the trocar), which can be placed near the patient.

### 3.1. The hybrid parallel robot kinematic description

Figure 2 presents the kinematic scheme of the modular parallel robot, consisting of one parallel mechanism (PM) and one spherical module (SM), and a surgical instrument (SI). The following mechanism topology can be identified:
1) PM consists of three kinematic chains:
   a. The first kinematic chain (KC₁) is type (2-$\underline{P}$R-R)-2P (Figure 2.b - blue), with two active prismatic joints $q_i, i=1,2$ (actuated along an axis parallel to OY), four passive revolute joints $R_{iP}, i=\overline{4,7}$ (with rotation axes parallel to OZ) and three passive prismatic joints $P_{iP}, i=\overline{1,3}$ (with orthogonal translation axes, $P_{3P}$ parallel to OY, and $P_{iP}, i=1,2$ parallel to OX); the links $l_3$ connect $R_{iP}, i=6,7$ with $R_{iP}, i=4,5$, respectively. KC₁ represents a 2 DOFs planar active joint that moves the element $P_{3P}$ in a 2T motion in the Cartesian plane XOY.

b. The second kinematic chain (KC2) is type 2-PUR (Figure 2.b - red), with the same active joints as KC1 ($q_i, i = 1,2$), two passive universal joints $u_{ip}, i = 1,2$, and two passive revolute joints $R_{ip}, i = 8,9$, connected through the links $l_1$. KC2 represents a 3 DOFs spatial mechanism with 2T-1R motion where the rotation motions are passive around the common rotation axis of the universal joints $u_{ip}, i = 1,2$.

c. The third kinematic chain (KC3) is type RRR (Figure 2.b - green), with one active revolute joint $q_3$ and two passive revolute joints $R_{ip}, i = 3,8$ (actuated around an axis parallel to OY), connected through the links $l_2$. KC3 is connected with KC1 through $P_{3P}$ and with KC2 through the cardan element $c$.

d. The cardan element $c$ is also connected to the passive cardan joint composed of the revolute joints $R_{ip}, i = 1,2$ and the surgical instrument ins mounted such that it intersects both rotation axes (in point $P_1$) of the revolute joints $R_{ip}, i = 1,2$.

2) SM consists of 5 passive revolute joints $R_{is}, i = \overline{1,5}$ with axes intersecting in the RCM point, and one passive cylindrical joint $C_{1s}$; the surgical instrument is allowed to rotate around the axis of $C_{1s}$ and slide along the same axis. The insertion depth of the surgical instrument of length $l$ is $l_{ins}$, with respect to the RCM point, and the rotation of the surgical instrument around the axis of $C_{1s}$ is achieved through $q_4$.

3) SI is mounted using the cardan joint composed of $R_{ip}, i = 1,2$ (within PM) and the cylindrical joint $C_{1s}$ (within SM). Consequently, SI passes through the RCM point. Two points are defined, E($X_E$, $Y_E$, $Z_E$) at the tip of the instrument, and P($X_P$, $Y_P$, $Z_P$) the point where SI is mounted to PM.

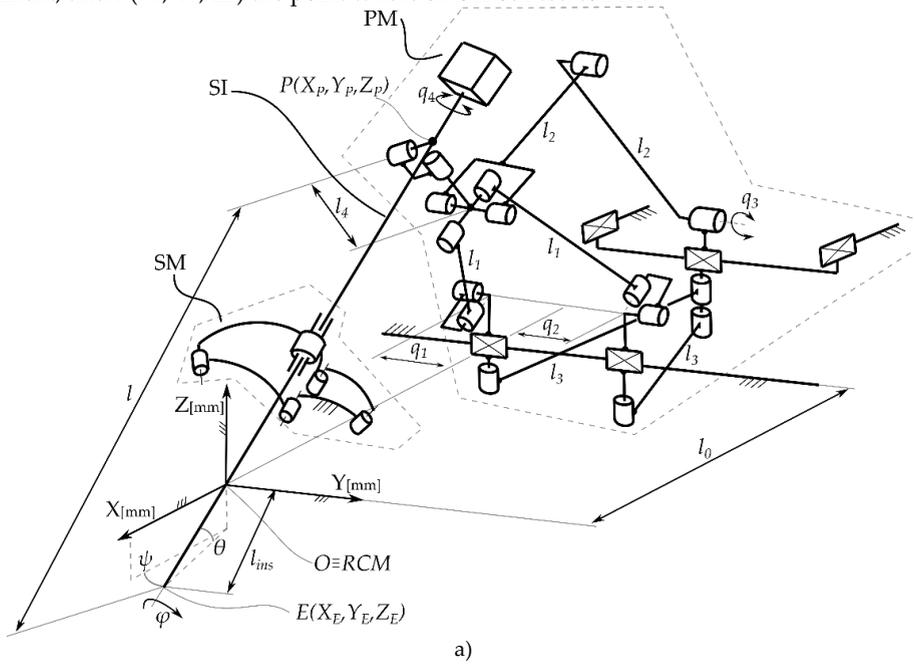

a)

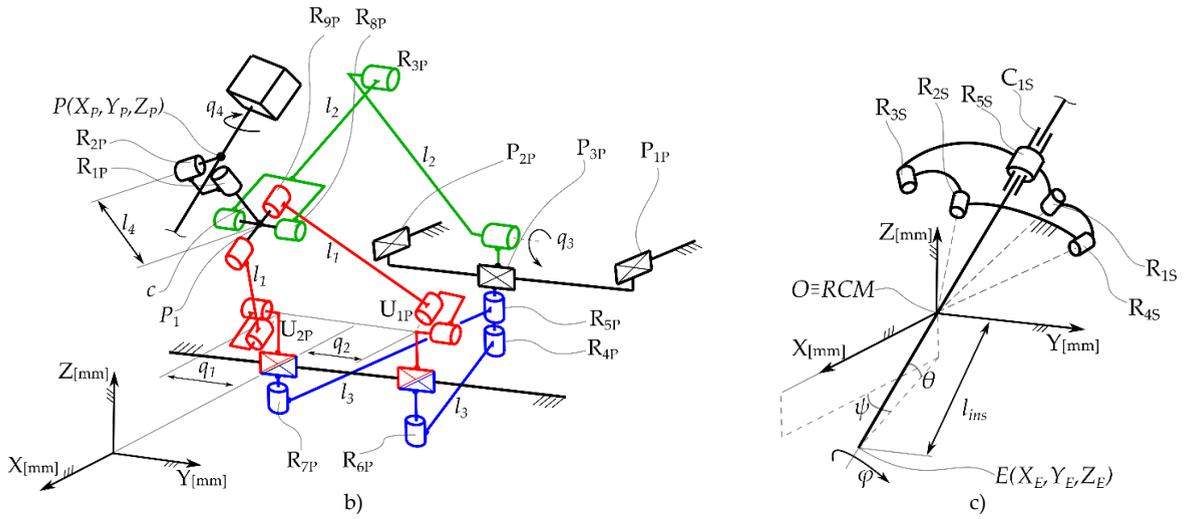

**Figure 2.** The kinematic scheme of the 4 DOFs hybrid parallel robot for pancreatic MIS: a) the complete modular hybrid parallel robot containing a parallel module (PM), a spherical module (SM), and a surgical instrument (SI); b) detail on the parallel module (PM); c) detail on the spherical module (SM).

### 3.2. Mechanism synthesis of the hybrid parallel robot

For the mechanism synthesis the following formula is used [38]:

$$M = (6-F) \cdot N - (5-F) \cdot C_5 - (4-F) \cdot C_4 - (3-F) \cdot C_3 - (2-F) \cdot C_2 - (1-F) \cdot C_1 \qquad (16)$$

which is a modification of Gruebler formula. Eq. (16) is useful for determining the mobility (number of independent DOFs) of parallel mechanisms. $M$ represents the mobility of the mechanism, $N$ the number of components in motion, $F$ is the family of the mechanism (defined as the number of the common constraints of the moving parts [38]), $C_i, i = \overline{1,5}$ represents the number of joints of class $i$ (class $i$ joints are defined as joints with $6-i$ DOFs [38]), and the factors $(i-F), i = \overline{1,6}$ must be strictly positive.

Computing the mobility of the hybrid robot, using Eq. (16), requires some simplifications to account for the modular design and redundancies. Two steps are considered in the calculation, both dependent on the Cardan element $c$ and its characteristic point $P_1$ (Figure 2.b):

1) The mobility of the Cardan element $c$ is computed based on the kinematic chains of PM. The following assumptions are made:
   a. $KC_1$ is defined as a class $C_4$ joint (having two degrees of freedom) that moves $P_{3P}$ (which is considered the first moving component).
   b. $KC_2$ is also defined as a class $C_4$ joint that moves the element $c$ (the second moving component), despite being a 3 DOFs kinematic chain; the motion produced by asynchronous actuation of $q_i, i = 1,2$ must be disregarded in the analysis since it is dependent on 1 DOF from $KC_1$; the motion produced by synchronous actuation of $q_i, i = 1,2$ cannot be disregarded (otherwise the mechanism is planar).
   c. $KC_3$ is defined as a class $C_3$ joint with 3 DOFs that also moves the element $c$.

**Table 1.** Mobility analysis of the PM components.

| Element | $T_X$ | $T_Y$ | $T_Z$ | $R_X$ | $R_Y$ | $R_Z$ |
|---|---|---|---|---|---|---|
| $P_{3P}$ | + | + | − | − | − | − |
| $c$ | + | − | + | − | + | − |

A mobility analysis for PM is shown in Table 1. Based on Table 1 and the above assumptions the numerical values of the parameters in Eq. (16) are: the family of the mechanism is $F = 2$ (since there are two rotation DOFs suppressed) and $N = 2$, $C_5 = 0$, $C_4 = 2$, $C_3 = 1$ ($C_2, C_1$ are not defined in this case since their factors are not strictly positive). Consequently, the mobility of PM (up to the Cardan element $c$) is $M = 3$. The motion type of PM, namely the Cardan element *c*, can be deduced by analyzing KC$_2$; it is clear that the mechanism has rotation around OY axis (the coinciding rotation axis of the two Cardan joints U$_{1P}$, U$_{2P}$) and two translations, one along OY axis and the second one along an axis orthogonal to both OY and the second axes of rotation of U$_{1P}$, U$_{2P}$. Consequently, the motion type of PM (up to the Cardan element *c*) is defined as R-2T (one rotation followed by two translations).

2) The mobility of the end-effector (surgical tool tip with characteristic point *E*) is computed based on the mobility of *c*. The following assumptions are made:
   a. The Cardan element *c* is guided by a class $c_3$ joint (the output motion of PM up to element *c*).
   b. The cylindrical element C$_{1S}$ (Figure 2.c) is guided by SM which is considered a class $c_2$ joint.
   c. SI (Figure 2.a) is the third moving element and it is constrained by SM through C$_{1S}$ and by PM through the Cardan joint composed by R$_{1P}$, R$_{2P}$ supported by *c* (Figure 2.b), and the active joint $q_4$.

**Table 2.** Mobility analysis of the end-effector.

| Element | $T_X$ | $T_Y$ | $T_Z$ | $R_X$ | $R_Y$ | $R_Z$ |
|---------|-------|-------|-------|-------|-------|-------|
| $C_{1S}$ | + | + | − | − | − | − |
| $c$ | + | − | + | − | + | − |
| SI | + | + | + | + | + | + |

A mobility analysis for the end-effector is shown in Table 2. The parameters required for the mobility computation (Eq. 16) are: $F = 0$, $N = 3$, $C_5 = 1$, $C_4 = 1$, $C_3 = 1$, $C_2 = 1$, $C_1 = 0$. The mobility of the end-effector is $M = 4$. The motion type of the end-effector can be described with respect to the RCM; since SI is constrained by SM through C$_{1S}$, it follows that the motion type of the end-effect is 3R-1T.

### 3.3. A simplified parameterization for the hybrid parallel robot

It is possible to define a simpler kinematic representation (with the same output motion) of the modular mechanism shown in Figure 2.

**Remark 1**: For the proposed design of the modular parallel robot, the position of the RCM point is always known, and is defined as the origin of the fixed coordinate system *OXYZ*.

**Remark 2**: The type of motion performed by the surgical instrument with respect to the RCM point is 3R + 1T. The RCM parameterization is defined as a rotation matrix $\mathbf{R}_{RCM} \in SO(3)$ with $\psi$, $\theta$, $\phi$, parameters following a $Z - Y - X$ rotation sequence, and a position vector **E** which a translation on the local *OX* axis by $l_{ins}$ (the insertion length), as defined in Eq. (17). Furthermore, the rotation angle due to the active revolute joint $q_4$ is defined such that is equivalent with $\phi$.

$$\mathbf{R}_{RCM} = \mathbf{R}_Z(\psi) \cdot \mathbf{R}_Y(\theta) \cdot \mathbf{R}_X(\varphi), \quad (\varphi = q_4),$$

$$\mathbf{E} = \mathbf{R}_{RCM} \cdot [l_{ins}, 0, 0]^T,$$

$$\mathbf{R}_Z(\psi) = \begin{bmatrix} \cos(\psi) & -\sin(\psi) & 0 \\ \sin(\psi) & \cos(\psi) & 0 \\ 0 & 0 & 1 \end{bmatrix}, \mathbf{R}_Y(\theta) = \begin{bmatrix} \cos(\theta) & 0 & \sin(\theta) \\ 0 & 1 & 0 \\ -\sin(\theta) & 0 & \cos(\theta) \end{bmatrix}, \mathbf{R}_X(\varphi) = \begin{bmatrix} \cos(\varphi) & -\sin(\varphi) & 0 \\ \sin(\varphi) & \cos(\varphi) & 0 \\ 0 & 0 & 1 \end{bmatrix}. \quad (17)$$

**Remark 3**: The points $E(X_E, Y_E, Z_E)$ (the tip of SI), $P(X_P, Y_P, Z_P)$ (where SI is mounted into PM) and RCM are always colinear.

**Remark 4**: The active joint parameter $q_4 = \phi$ (Figure 2) is decoupled from the rest of the parallel mechanism PM and the rotation axis of $q_4$ passes through a point $P(X_P, Y_P, Z_P)$ which in turn is the intersection of the rotation axes of the $R_{P1}$ and $R_{P2}$ joints. Point *P* is defined to significantly simplify the kinematic models of the modular hybrid parallel robot for pancreatic MIS.

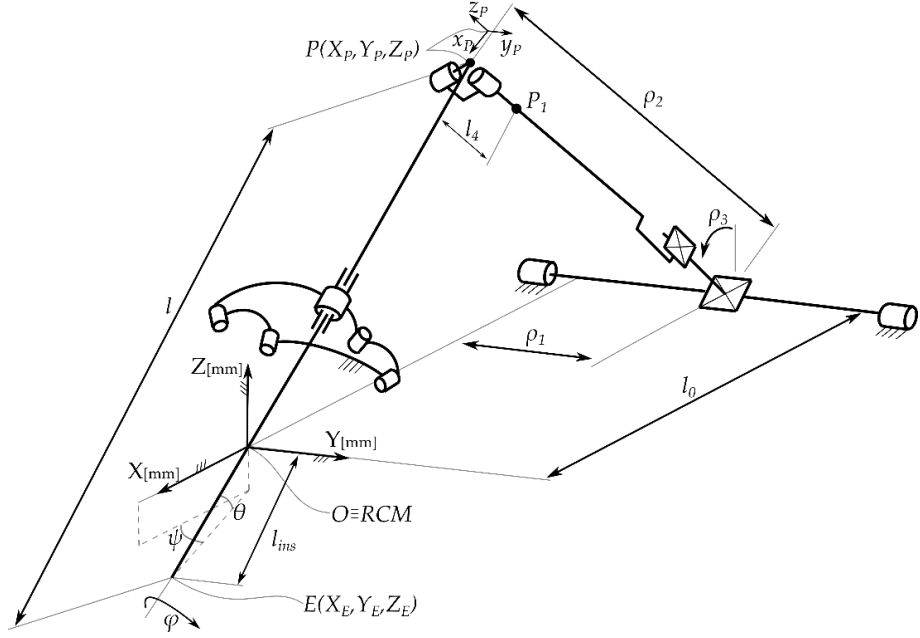

**Figure 3.** The kinematic scheme of the simplified parametrization as an equivalent 3-DOF surgical robot.

Accounting for Remarks 1 – 4, the kinematic analysis of the 4-DOF mechanism (Figure 2) can be simplified using a serial representation (Figure 3), where the active joint $q_4$ was excluded which in turn simplifies the kinematic models. As seen, the active prismatic joints $q_1$ and $q_2$ have been replaced by the active prismatic joints $\rho_1$ and $\rho_2$ which perform the same motion. Moreover, $q_3$ has been replaced by $\rho_3$.

Considering the kinematic scheme in Figure 3, the serial representation <u>RPP</u> (up to point $P_1$) produces the same motion (R-2T) as the parallel mechanism.

Eq. (18) shows the input-output relations that describe the mathematical connection between the two sets of parameters $q_i$, $i = \overline{1,3}$ (Figure 2) and $\rho_i$, $i = \overline{1,3}$ (Figure 3). The derivation of these equations is shown in Appendix B.

$$IO_{\rho\_q} : \begin{cases} \rho_1 - \frac{1}{2}(q_1 + q_2) = 0 \\ \left(\frac{1}{2}q_1 + \frac{1}{2}q_2\right)^2 + (\rho_2 - l_4)^2 - l_1^2 = 0 \\ \left(l_3' - l_2\sin(q_3) + l_1'\sin(\rho_3)\right)^2 + \left(l_2\cos(q_3) - l_1'\cos(\rho_3)\right)^2 - l_2^2 = 0 \end{cases} \quad (18)$$

$$l_1' = \sqrt{l_1^2 - \left(\tfrac{1}{2}q_1 + \tfrac{1}{2}q_2\right)^2},\ l_3' = \sqrt{l_3^2 - \left(\tfrac{1}{2}q_1 + \tfrac{1}{2}q_2\right)^2},$$

that can be solved either for $q_i$, $i = \overline{1,3}$ (yielding 4 solutions) or for $\rho_i$, $i = \overline{1,3}$ (yielding 4 solutions). Appendix C1 shows numerical examples regarding the mapping between $q_i$, $i = \overline{1,3}$ and $\rho_i$, $i = \overline{1,3}$ parameters.

## 4. Kinematic modeling

The section focuses on developing the kinematic models for the modular hybrid parallel robot for pancreatic surgery up to the jerk field; it is possible to compute the higher acceleration up to an arbitrary degree, the jerk field is chosen since standard command and control hardware (e.g., servomotors, drivers) are jerk limited. Although both the forward and inverse kinematic algorithms were developed and used in the analysis (Section 5), only the inverse kinematic algorithms are reported (as algorithms) in this work. All kinematic models yield closed form symbolic solutions, and the forward kinematics is reciprocal to the inverse kinematics. Furthermore, the mapping between the real active join parameters $q_i$, $i = \overline{1,3}$ (Figure 2) and the simplified active joint parameters $\rho_i$, $i = \overline{1,3}$ (Figure 3) will be used (identically) in all kinematic algorithms.

## 4.1. Preliminary symbolic computation and required evaluation technique

### 4.1.1. Higher-order kinematics for the mapping between the real active joint parameters and the simplified ones

As previously stated, Eq. (18) can be solved for both $q_i$, $i = \overline{1,3}$ or $\rho_i$, $i = \overline{1,3}$ parameters yielding the inverse and forward kinematics (for displacement), respectively. To obtain the higher-order kinematics the following equation is used (as in Eq. 2):

$$\mathbf{A}_{\rho\_q} \cdot \dot{\mathbf{X}}_{\rho} + \mathbf{B}_{\rho\_q} \cdot \dot{\mathbf{Q}}_{q} = 0, \tag{19}$$

where $\mathbf{A}_{\rho\_q}$ and $\mathbf{B}_{\rho\_q}$ are the Jacobian matrices computed using Eq. (17), by differentiation with respect to $\mathbf{X}_{\rho} = [\rho_1, \rho_2, \rho_3]^T$ and $\mathbf{Q}_q = [q_1, q_2, q_3]^T$, respectively. Assuming that $\mathbf{A}_{\rho\_q}$ and $\mathbf{B}_{\rho\_q}$ are nonsingular, Eq. (19) can be rewritten as:

$$\dot{\mathbf{Q}}_q = \mathbf{J}_{\rho\_q} \cdot \dot{\mathbf{X}}_{\rho}, \quad \mathbf{J}_{\rho\_q} = -\mathbf{B}_{\rho\_q}^{-1} \cdot \mathbf{A}_{\rho\_q}, \tag{20.a}$$

$$\dot{\mathbf{X}}_{\rho} = \mathbf{J}_{\rho\_q}^{-1} \cdot \dot{\mathbf{Q}}_q, \tag{20.b}$$

where Eq. (20.a) yields the inverse kinematics for velocities and Eq. (20.b) yields the forward kinematics, respectively. To compute the higher-order kinematics for the mapping between $q_i$, $i = \overline{1,3}$ and $\rho_i$, $i = \overline{1,3}$, Eqs. (20.a) and (20.b) are either differentiated or multidualized. *Algorithm 1 (IK – map q-ρ)* shows how to develop an inverse kinematic function for the aforementioned mapping, which includes displacement and the higher velocities up to the jerk field. The algorithm is straightforward, but its purpose is to highlight a key difference in implementation versus the multidual approach.

*Algorithm 1 (IK – map q-ρ)*

**0. Inputs**: $\mathbf{X}_{\rho} = [\rho_1, \rho_2, \rho_3]^T, \dot{\mathbf{X}}_{\rho}, \ddot{\mathbf{X}}_{\rho}, \dddot{\mathbf{X}}_{\rho}$

**1. Compute**: $\mathbf{Q}_q = [q_1, q_2, q_3]^T$ using Eq. (18) and $\dot{\mathbf{Q}}_q$, $\ddot{\mathbf{Q}}_q$, $\dddot{\mathbf{Q}}_q$ using Eq. (20.a) and its derivatives:

$$\dot{\mathbf{Q}}_q = \mathbf{J}_{\rho\_q} \cdot \dot{\mathbf{X}}_{\rho},\quad \ddot{\mathbf{Q}}_q = \dot{\mathbf{J}}_{\rho\_q} \cdot \dot{\mathbf{X}}_{\rho} + \mathbf{J}_{\rho\_q} \cdot \ddot{\mathbf{X}}_{\rho},\quad \dddot{\mathbf{Q}}_q = \ddot{\mathbf{J}}_{\rho\_q} \cdot \dot{\mathbf{X}}_{\rho} + 2 \cdot \dot{\mathbf{J}}_{\rho\_q} \cdot \ddot{\mathbf{X}}_{\rho} + \mathbf{J}_{\rho\_q} \cdot \dddot{\mathbf{X}}_{\rho}$$

**2. Outputs**: $\mathbf{Q}_q = [q_1, q_2, q_3]^T, \dot{\mathbf{Q}}_q, \ddot{\mathbf{Q}}_q, \dddot{\mathbf{Q}}_q$

### 4.1.2. Higher-order kinematics technique to implement multidual algebra

Since multidual algebra computes simultaneously the higher-order kinematics, encoded in the multidual quantities (as coefficients of $\varepsilon^i / i!$, $i = \overline{1,n}$), a technique that first evaluates the symbolic expressions for the input parameters (geometric parameters, displacement, velocities, etc.), and then extract the desired coefficients for the higher-order kinematics is desired. This can be achieved by implementing function handles (in MATLAB) or pointers (in C), and the following steps are required for the logical flow if the computation (the example is given up to the jerk field):

1. Compute the multidual parameter $\overset{(}{s}$ for $\varepsilon^i \neq 0$, $i = \overline{1,2}$, $\varepsilon^3 = 0$
2. Evaluate $\overset{(}{s}$ for all input parameters (create function handle or pointer to a function that holds the $\overset{(}{s}$ parameter symbolically) to obtain $\overset{(}{s}_0$ (which simplifies as a polynomial in $\varepsilon$)
3. Commute the values of $\varepsilon^i$ between 0 and 1 to obtain the desired higher-order kinematic field: a) for velocity evaluate $\overset{(}{s}_0$ for $[\varepsilon^0 = 1, \varepsilon^1 = 0, \varepsilon^2 = 0]$; b) for acceleration evaluate $\overset{(}{s}_0$ for $[\varepsilon^0 = 0, \varepsilon^1 = 1, \varepsilon^2 = 0]$; c) for jerk evaluate $\overset{(}{s}_0$ for $[\varepsilon^0 = 0, \varepsilon^1 = 0, \varepsilon^2 = 1]$

The above example is designed for the Jacobian matrix multidualization; if other formalism is used (homogeneous matrix, or dual quaternion) then the powers of $\varepsilon$ must be changed accordingly.

*Algorithm 2 (IK – multidual q-ρ)* shows how to implement multidual algebra in the inverse kinematics for the mapping between $q_i$, $i = \overline{1,3}$ and $\rho_i$, $i = \overline{1,3}$. The tradeoff between classical differentiation implementation and the multidual implementation is that multidual algebra evaluates all the kinematic fields simultaneously (as opposed to differentiation where the evaluation is sequential) but requires the computation of $\varepsilon^i$ between 0 and 1 to get the desired kinematic field. It turns out that the latter approach can speed up the execution time, at least for the proposed parallel robot in this work.

---

*Algorithm 2 (IK – multidual q-ρ)*

**0. Inputs**: $\mathbf{X}_\rho = [\rho_1, \rho_2, \rho_3]^T, \overset{(}{\mathbf{X}}_\rho$

**1. Compute**: $\mathbf{Q}_q = [q_1, q_2, q_3]^T$ using Eq. (18) and $\overset{(}{\mathbf{Q}}_q$ :

$$\overset{(}{\mathbf{Q}}_q = \overset{(}{\mathbf{J}}_{\rho\_q} \cdot \overset{(}{\mathbf{X}}_\rho$$

**2. Set** $\varepsilon^k \neq 0, \varepsilon^n = 0, n > 2, k = 1, 2$ and

**Compute**: $\overset{\&}{\mathbf{Q}}_q$ (evaluate $\overset{(}{\mathbf{Q}}_q$ for $[\varepsilon^0 = 1, \varepsilon^1 = 0, \varepsilon^2 = 0]$ )

$\overset{\&\&}{\mathbf{Q}}_q$ (evaluate $\overset{(}{\mathbf{Q}}_q$ for $[\varepsilon^0 = 0, \varepsilon^1 = 1, \varepsilon^2 = 0]$ )

$\overset{\&\&\&}{\mathbf{Q}}_q$ (evaluate $\overset{(}{\mathbf{Q}}_q$ for $[\varepsilon^0 = 0, \varepsilon^1 = 0, \varepsilon^2 = 1]$ )

**3. Outputs**: $\mathbf{Q}_q = [q_1, q_2, q_3]^T, \overset{\&}{\mathbf{Q}}_q, \overset{\&\&}{\mathbf{Q}}_q, \overset{\&\&\&}{\mathbf{Q}}_q$

---

### 4.2. Kinematic modeling of the hybrid parallel robot (geometric approach)

The input-output relations that connect $\rho_i, i = \overline{1,3}$ with $P(X_P, Y_P, Z_P)$ are (Figure 3):

$$IO_{P\_\rho} : \begin{cases} X_P + l_0 - \rho_2 \sin(\rho_3) = 0 \\ Y_P - \rho_1 = 0 \\ Z_P - \rho_2 \cos(\rho_3) = 0 \end{cases} \quad (21)$$

which can be solved either for $X_P, Y_P, Z_P$ (yielding a unique solution) or for $\rho_i, i = \overline{1,3}$ (yielding 2 solutions).

The position of point $P(X_P, Y_P, Z_P)$ is mathematically related to the RCM parameters $\psi, \theta, l_{ins}$ through:

$$IO_{RCM\_P} : \begin{cases} X_P - (-l + l_{ins})\cos(\psi)\cos(\theta) = 0 \\ Y_P - (-l + l_{ins})\sin(\psi)\cos(\theta) = 0 \\ Z_P + (-l + l_{ins})\sin(\theta) = 0 \end{cases} \quad (22)$$

which can be solved either for $X_P, Y_P, Z_P$ (yielding unique solution) or for $\psi, \theta, l_{ins}$ (yielding 4 solutions).

Lastly, the position of point $E(X_E, Y_E, Z_E)$ (the tip of the instrument) is mathematically related with $\psi, \theta, l_{ins}$ through:

$$IO_{E\_RCM} : \begin{cases} X_E - l_{ins} \cos(\psi)\cos(\theta) = 0 \\ Y_E - l_{ins} \sin(\psi)\cos(\theta) = 0 \\ Z_E + l_{ins} \sin(\theta) = 0 \end{cases} \quad (23)$$

which can be solved either for $X_E, Y_E, Z_E$ (yielding unique solution) or for $\psi, \theta, l_{ins}$ (yielding 4 solutions). All four sets of input-output equations shown in this sub-section can be solved for the inverse and forward kinematic models yielding closed form solutions. Appendix C2 shows numerical examples regarding the mapping between $X_E, Y_E, Z_E$ and $\rho_i, i = \overline{1,3}$ parameters.

#### 4.2.1. Higher-order kinematics using the kinematic Jacobians and differentiation

The following matrix equations are defined for the higher-order kinematics for the modular parallel robot:

$$\dot{\mathbf{Q}}_\rho = \mathbf{J}_{P\_\rho} \cdot \dot{\mathbf{X}}_P, \mathbf{J}_{P\_\rho} = -\mathbf{B}_{P\_\rho}^{-1} \cdot \mathbf{A}_{P\_\rho}, \quad (24.\text{a})$$

$$\dot{\mathbf{X}}_P = \mathbf{J}_{P\_\rho}^{-1} \cdot \dot{\mathbf{Q}}_\rho, \quad (24.\text{b})$$

$$\dot{\mathbf{Q}}_P = \mathbf{J}_{RCM\_P} \cdot \dot{\mathbf{X}}_R, \mathbf{J}_{RCM\_P} = -\mathbf{B}_{RCM\_P}^{-1} \cdot \mathbf{A}_{RCM\_P}, \quad (25.\text{a})$$

$$\dot{\mathbf{X}}_R = \mathbf{J}_{RCM\_P}^{-1} \cdot \dot{\mathbf{Q}}_P, \quad (25.\text{b})$$

$$\dot{\mathbf{Q}}_R = \mathbf{J}_{RCM\_E} \cdot \dot{\mathbf{X}}_E, \mathbf{J}_{RCM\_E} = -\mathbf{B}_{RCM\_E}^{-1} \cdot \mathbf{A}_{RCM\_E}, \quad (26.\text{a})$$

$$\dot{\mathbf{X}}_E = \mathbf{J}_{RCM\_E}^{-1} \cdot \dot{\mathbf{Q}}_R, \quad (26.\text{b})$$

where $\mathbf{A}_{P\_\rho}$ and $\mathbf{B}_{P\_\rho}$ are the Jacobian matrices computed using Eq. (21), by differentiation with respect to $\mathbf{X}_P = [X_P, Y_P, Z_P]^T$ and $\mathbf{Q}_\rho = [\rho_1, \rho_2, \rho_3]^T$, respectively; $\mathbf{A}_{RCM\_P}$ and $\mathbf{B}_{RCM\_P}$ are the Jacobian matrices computed from Eq. (22), by differentiation with respect to $\mathbf{X}_R = [\psi, \theta, l_{ins}]^T$ and $\mathbf{Q}_P = [X_P, Y_P, Z_P]^T$, respectively; finally, $\mathbf{A}_{RCM\_E}$ and $\mathbf{B}_{RCM\_E}$ are the Jacobian matrices derived from Eq. (23) by differentiating with respect to $\mathbf{X}_E = [\mathbf{X}_E, \mathbf{Y}_E, \mathbf{Z}_E]^T$ and $\mathbf{Q}_R = [\psi, \theta, l_{ins}]^T$, respectively. Eqs. (24.a), (25.a), (26.a) are used for the inverse kinematics, whereas Eqs. (24.b), (25.b), (26.b) are used for forward kinematics.

It should be noted that the kinematics of the proposed modular parallel robot is computed sequentially solving of the velocity vectors and back substitution. *Algorithm 3 (Inverse Kinematics – Geometric Approach)* shows how to compute the *Inverse Kinematics* for the modular parallel robot using the kinematic Jacobians.

*Algorithm 3 (Inverse Kinematics – Geometric Approach)*

---

**0. Inputs**: instrument tip parameters: $\mathbf{X}_E = [X_E, Y_E, Z_E]^T, \dot{\mathbf{X}}_E, \ddot{\mathbf{X}}_E, \dddot{\mathbf{X}}_E$

1. **Compute**: $\mathbf{Q}_R = [\psi, \theta, l_{ins}]^T$ using Eq. (23) and $\dot{\mathbf{Q}}_R, \ddot{\mathbf{Q}}_R, \dddot{\mathbf{Q}}_R$ using Eq. (26.a) and its derivatives:

$$\dot{\mathbf{Q}}_R = \mathbf{J}_{RCM\_E} \cdot \dot{\mathbf{X}}_E, \ddot{\mathbf{Q}}_R = \dot{\mathbf{J}}_{RCM\_E} \cdot \dot{\mathbf{X}}_E + \mathbf{J}_{RCM\_E} \cdot \ddot{\mathbf{X}}_E, \dddot{\mathbf{Q}}_R = \ddot{\mathbf{J}}_{RCM\_E} \cdot \dot{\mathbf{X}}_E + 2 \cdot \dot{\mathbf{J}}_{RCM\_E} \cdot \ddot{\mathbf{X}}_E + \mathbf{J}_{RCM\_E} \cdot \dddot{\mathbf{X}}_E$$

2. **Set**: $\mathbf{X}_R = \mathbf{Q}_R, \dot{\mathbf{X}}_R = \dot{\mathbf{Q}}_R, \ddot{\mathbf{X}}_R = \ddot{\mathbf{Q}}_R, \dddot{\mathbf{X}}_R = \dddot{\mathbf{Q}}_R$ and

**Compute**: $\mathbf{Q}_P = [X_P, Y_P, Z_P]^T$ using Eq. (22) $\dot{\mathbf{Q}}_P, \ddot{\mathbf{Q}}_P, \dddot{\mathbf{Q}}_P$ using Eq. (25.a) and its derivatives:

$$\dot{\mathbf{Q}}_P = \mathbf{J}_{RCM\_\rho} \cdot \dot{\mathbf{X}}_R, \ddot{\mathbf{Q}}_P = \dot{\mathbf{J}}_{RCM\_\rho} \cdot \dot{\mathbf{X}}_R + \mathbf{J}_{RCM\_\rho} \cdot \ddot{\mathbf{X}}_R, \dddot{\mathbf{Q}}_P = \ddot{\mathbf{J}}_{RCM\_\rho} \cdot \dot{\mathbf{X}}_R + 2 \cdot \dot{\mathbf{J}}_{RCM\_\rho} \cdot \ddot{\mathbf{X}}_R + \mathbf{J}_{RCM\_\rho} \cdot \dddot{\mathbf{X}}_R$$

3. **Set**: $\mathbf{X}_P = \mathbf{Q}_P, \dot{\mathbf{X}}_P = \dot{\mathbf{Q}}_P, \ddot{\mathbf{X}}_P = \ddot{\mathbf{Q}}_P, \dddot{\mathbf{X}}_P = \dddot{\mathbf{Q}}_P$, and

**Compute**: $\mathbf{Q}_\rho = [\rho_1, \rho_2, \rho_3]^T$ using (21) and $\dot{\mathbf{Q}}_\rho, \ddot{\mathbf{Q}}_\rho, \dddot{\mathbf{Q}}_\rho$ using Eq. (24.a) and its derivatives:

$$\dot{\mathbf{Q}}_\rho = \mathbf{J}_{P\_\rho} \cdot \dot{\mathbf{X}}_P, \ddot{\mathbf{Q}}_\rho = \dot{\mathbf{J}}_{P\_\rho} \cdot \dot{\mathbf{X}}_P + \mathbf{J}_{P\_\rho} \cdot \ddot{\mathbf{X}}_P, \dddot{\mathbf{Q}}_\rho = \ddot{\mathbf{J}}_{P\_\rho} \cdot \dot{\mathbf{X}}_P + 2 \cdot \dot{\mathbf{J}}_{P\_\rho} \cdot \ddot{\mathbf{X}}_P + \mathbf{J}_{P\_\rho} \cdot \dddot{\mathbf{X}}_P$$

4. **Outputs**: $\mathbf{Q}_\rho = [\rho_1, \rho_2, \rho_3]^T, \dot{\mathbf{Q}}_\rho, \ddot{\mathbf{Q}}_\rho, \dddot{\mathbf{Q}}_\rho$

---

To determine the required higher-order kinematics for the real parallel robot, the outputs of *Algorithm 3* are passed as inputs in *Algorithm 1*.

*4.2.2. Higher-order kinematics using the multidual kinematic Jacobians*

The multidual approach also uses Eqs. (24.a), (25.a), (26.a), but instead of differentiation, dualization is applied. *Algorithm 4 (Inverse Kinematics – Multidual Approach)* shows how to compute the *Inverse Kinematics* for the modular parallel robot using the multidualization of the kinematic Jacobians.

*Algorithm 4 (Inverse Kinematics – Multidual Approach)*

**0. Inputs**: instrument tip parameters: $\mathbf{X}_E = [X_E, Y_E, Z_E]^T$, $\overset{(}{\mathbf{X}}_E = \dot{\mathbf{X}}_E + \varepsilon \ddot{\mathbf{X}}_E + \frac{\varepsilon^2}{2}\dddot{\mathbf{X}}_E$

**1. Compute**: $\mathbf{Q}_R = [\psi, \theta, l_{ins}]^T$ using Eq. (23) and $\overset{(}{\mathbf{Q}}_R$ using Eq. (26.a):

$$\overset{(}{\mathbf{Q}}_R = \overset{(}{\mathbf{J}}_{RCM\_E} \cdot \overset{(}{\mathbf{X}}_E$$

**2. Set:** $\mathbf{x}_R = \mathbf{Q}_R$, $\overset{(}{\mathbf{X}}_R = \overset{(}{\mathbf{Q}}_R$ and

**Compute**: $\mathbf{Q}_P = [X_P, Y_P, Z_P]^T$ using Eq. (22) and $\overset{(}{\mathbf{Q}}_P$ using Eq. (25.a):

$$\overset{(}{\mathbf{Q}}_P = \overset{(}{\mathbf{J}}_{RCM\_\rho} \cdot \overset{(}{\mathbf{X}}_R$$

**3. Set:** $\mathbf{X}_P = \mathbf{Q}_P$, $\overset{(}{\mathbf{X}}_P = \overset{(}{\mathbf{Q}}_P$, and

**Compute**: $\mathbf{Q}_\rho = [\rho_1, \rho_2, \rho_3]^T$ using Eq. (21) and $\overset{(}{\mathbf{Q}}_\rho$ using Eq. (24.a):

$$\overset{(}{\mathbf{Q}}_\rho = \overset{(}{\mathbf{J}}_{P\_\rho} \cdot \overset{(}{\mathbf{X}}_P$$

**4. Set** $\varepsilon^k \neq 0$, $\varepsilon^n = 0$, $n > 2$, $k = 1, 2$, and

**Compute**: $\dot{\mathbf{Q}}_\rho$ (evaluate $\overset{(}{\mathbf{Q}}_\rho$ for $[\varepsilon^0 = 1, \varepsilon^1 = 0, \varepsilon^2 = 0]$)

$\ddot{\mathbf{Q}}_\rho$ (evaluate $\overset{(}{\mathbf{Q}}_\rho$ for $[\varepsilon^0 = 0, \varepsilon^1 = 1, \varepsilon^2 = 0]$)

$\dddot{\mathbf{Q}}_\rho$ (evaluate $\overset{(}{\mathbf{Q}}_\rho$ for $[\varepsilon^0 = 0, \varepsilon^1 = 0, \varepsilon^2 = 1]$)

**5. Outputs**: $\mathbf{Q}_\rho = [\rho_1, \rho_2, \rho_3]^T, \dot{\mathbf{Q}}_\rho, \ddot{\mathbf{Q}}_\rho, \dddot{\mathbf{Q}}_\rho$

*4.3. Kinematic modeling of the hybrid parallel robot (homogeneous matrix approach)*

Point **P** and the frame attached to it simplifies the kinematic models for the homogeneous matrix approach since it represents straightforward mathematical connection with the $\rho_i, i = \overline{1,3}$ parameters, while also allowing to not consider the free motion parameters due to the revolute joints $R_{1P}$ and $R_{2P}$. It is natural to consider the position vector **P**$[X_P, Y_P, Z_P]^T$ with respect to both the RCM motion parameterization and the mechanism shown in Figure 3 and use simple transformations to change between **P**$[X_P, Y_P, Z_P]^T$ and **E**$[X_E, Y_E, Z_E]^T$. The following steps are followed to derive the kinematic models:

The following homogeneous transformation matrix describes the RCM parameterization:

$$\mathbf{M}_{P1} : \begin{bmatrix} \mathbf{R}_{RCM} & \mathbf{P}_1 \\ \mathbf{0} & 1 \end{bmatrix} = \begin{bmatrix} \mathbf{R}_Z(\psi) \cdot \mathbf{R}_Y(\theta) & \mathbf{R}_Z(\psi) \cdot \mathbf{R}_Y(\theta) \cdot \mathbf{d}_{L0} \\ \mathbf{0} & 1 \end{bmatrix},$$
$$\mathbf{P}_1 = [X_P, Y_P, Z_P]^T, \quad \mathbf{d}_{L0} = [-(l - l_{ins}), 0, 0]^T.$$
(27)

where $\mathbf{R}_Z$, and $\mathbf{R}_Y$ are rotation matrices in SO(3) with $\psi$, $\theta$ as angle parameters following a $Z - Y$ rotation convention. The position vector **P**$[X_P, Y_P, Z_P]^T$ is therefore:

$$\mathbf{P}_1 : \begin{bmatrix} X_P \\ Y_P \\ Z_P \end{bmatrix} = \begin{bmatrix} -(l - l_{ins})\cos(\psi)\cos(\theta) \\ -(l - l_{ins})\sin(\psi)\cos(\theta) \\ (l - l_{ins})\sin(\theta) \end{bmatrix}$$
(28)

On the other hand, the position vector **P**$[X_P, Y_P, Z_P]^T$ with respect to the simplified parameterization of the mechanism (Figure 3) is computed using:

$$\mathbf{M}_{P2} : \begin{bmatrix} \mathbf{R}_P & \mathbf{P}_2 \\ \mathbf{0} & 1 \end{bmatrix} = \begin{bmatrix} \mathbf{R}_Y(\rho_3) \cdot \mathbf{R}_Z(\alpha) \cdot \mathbf{R}_Y(\beta) & \mathbf{d}_{\rho 1} \cdot \mathbf{R}_Y(\rho_3) \cdot \mathbf{d}_{\rho 2} \\ \mathbf{0} & 1 \end{bmatrix},$$
$$\mathbf{P}_2 = [X_P, Y_P, Z_P]^T, \quad \mathbf{d}_{\rho 1} = [-l_0, \rho_1, 0]^T, \quad \mathbf{d}_{\rho 2} = [0, 0, \rho_2]^T.$$
(29)

It is assumed that $\alpha$ and $\beta$ are unknown angle parameters for the passive revolute joints $R_{P1}$, $R_{P2}$; however, as seen in Eq. (29) $\alpha$ and $\beta$ parameters do not contribute to the $\mathbf{P}_2$ vector. The vector $\mathbf{P}_2$ reads:

$$\mathbf{P}_2 : \begin{bmatrix} X_P \\ Y_P \\ Z_P \end{bmatrix} = \begin{bmatrix} \sin(\rho_3)\rho_2 - l_0 \\ \rho_1 \\ \cos(\rho_3)\rho_2 \end{bmatrix}. \tag{30}$$

Since the transformation matrices $\mathbf{M}_P$ and $\mathbf{M}_{P2}$ represent the same physical SE(3) displacement, therefore, setting $\mathbf{P}_1 = \mathbf{P}_2$ yields input-output equations that relate the RCM parameters $\psi$, $\theta$, $l_{ins}$ with the active joints parameters $\rho_i$, $i = \overline{1,3}$. Furthermore, RCM parameters are related to the tip of the surgical instrument parameters $X_E$, $Y_E$, $Z_E$ through:

$$\mathbf{M}_E : \begin{bmatrix} \mathbf{R}_{RCM} & \mathbf{E} \\ \mathbf{0} & 1 \end{bmatrix} = \mathbf{M}_{P1} \cdot \begin{bmatrix} \mathbf{I}_3 & \mathbf{d}_L \\ \mathbf{0} & 1 \end{bmatrix}, \tag{31}$$

$$\mathbf{E} = [X_E, Y_E, Z_E]^T, \quad \mathbf{d}_L = [L, 0, 0]^T.$$

Appendix C3 shows numerical examples regarding the mapping between $X_E$, $Y_E$, $Z_E$ and $\rho_i$, $i = \overline{1,3}$ parameters using the homogeneous matrix approach.

### 4.3.1. Higher-order kinematics using the derivatives of the homogeneous transformation matrix

To compute the higher-order kinematics based on the homogeneous matrices Eqs. (27), (29) are used as inputs for Eq. (9):

$$\mathbf{V}_{P1}^{(n)} = \mathbf{P}_1^{(n)} - \mathbf{R}_{RCM}^{(n)} \mathbf{R}_{RCM}^T \mathbf{P}_1 \tag{32}$$

$$\mathbf{V}_{P2}^{(n)} = \mathbf{P}_2^{(n)} - \mathbf{R}_P^{(n)} \mathbf{R}_P^T \mathbf{P}_2 \tag{33}$$

Since Eqs. (22), (24) represent the same Euclidean displacement, the higher-order kinematics is achieved through $\mathbf{V}_{P1} - \mathbf{V}_{P2} = \mathbf{0}$, which allows the computation of the higher kinematics of either the RCM parameters (forward kinematics) or the active joint parameters (inverse kinematics) in a sequential manner: n=1 yields velocities, n=2 yields accelerations and so on. *Algorithm 5 (Inverse Kinematics – homogeneous matrix)* shows how the formalism can be applied to obtain the inverse kinematics of the modular parallel robot.

---

*Algorithm 5 (Inverse Kinematics – homogeneous matrix)*

**0. Inputs**: instrument tip parameters: $\mathbf{X}_E = [X_E, Y_E, Z_E]^T, \dot{\mathbf{X}}_E, \ddot{\mathbf{X}}_E, \dddot{\mathbf{X}}_E$

**1. Compute**: $\mathbf{RCM} = [\psi, \theta, l_{ins}]^T$ using Eqs. (31), (27), (28) and

$\mathbf{Q}_\rho = [\rho_1, \rho_2, \rho_3]^T$ using $\mathbf{P}_1 - \mathbf{P}_2 = \mathbf{0}$

**2. Compute**: $\mathbf{v}_{P1}^{(n)}$, using Eqs. (32) for $n = \overline{1,3}$

**3. Compute**: $\mathbf{v}_{P2}^{(n)}$, using Eqs. (33) for $n = \overline{1,3}$

**4. Solve**: $\mathbf{v}_{P1}^{(n)} - \mathbf{v}_{P2}^{(n)} = \mathbf{0}$, for $\mathbf{Q}_\rho^{(n)} = [\rho_1^{(n)}, \rho_2^{(n)}, \rho_3^{(n)}]^T$, $n = \overline{1,3}$

**4. Outputs**: $\mathbf{Q}_\rho^{(n)} = [\rho_1^{(n)}, \rho_2^{(n)}, \rho_3^{(n)}]^T$, $n = \overline{0,3}$

---

### 4.3.2. Higher-order kinematics using the multidual homogeneous transformation matrix

To achieve the higher-order kinematics using the multidual homogeneous matrix formalism, the vectors $\mathbf{P}_1$, $\mathbf{P}_2$ (Eqs. 27, 29) are used. The kinematic models are derived sequentially.

$$\overset{(}{\mathbf{V}}_1: \begin{bmatrix} \overset{(}{X}_P \\ \overset{(}{Y}_P \\ \overset{(}{Z}_P \end{bmatrix} = \overset{(}{\mathbf{P}}_1 - \overset{(}{\mathbf{R}}_{RCM} \mathbf{R}_{RCM}^T \mathbf{P}_1 = \begin{bmatrix} (\overset{(}{l}_{ins} - l_{ins}) \cos(\overset{(}{\psi}) \cos(\overset{(}{\theta}) \\ (\overset{(}{l}_{ins} - l_{ins}) \sin(\overset{(}{\psi}) \cos(\overset{(}{\theta}) \\ -(\overset{(}{l}_{ins} - l_{ins}) \sin(\overset{(}{\theta}) \end{bmatrix}, \tag{34}$$

$$\overset{(}{\mathbf{V}}_2: \begin{bmatrix} \overset{(}{X}_P \\ \overset{(}{Y}_P \\ \overset{(}{Z}_P \end{bmatrix} = \overset{(}{\mathbf{P}}_2 - \overset{(}{\mathbf{R}}_P \mathbf{R}_P^T \mathbf{P}_2 = \begin{bmatrix} (l_0 \sin(\rho_3) - \rho_2 + \overset{(}{\rho}_2) \sin(\overset{(}{\rho}_3) + l_0 (\cos(\overset{(}{\rho}_3) \cos(\rho_3) - 1) \\ \overset{(}{\rho}_1 - \rho_1 \\ (l_0 \sin(\rho_3) - \rho_2 + \overset{(}{\rho}_2) \cos(\overset{(}{\rho}_3) - l_0 \cos(\overset{(}{\rho}_3) \sin(\rho_3) \end{bmatrix}. \tag{35}$$

Based on the collinearity of points E, RCM, P, the higher-order kinematics of **E** will have the same mathematical description as **P**, i.e., $[\overset{(}{X}_P, \overset{(}{Y}_P, \overset{(}{Z}_P]^T = [\overset{(}{X}_E, \overset{(}{Y}_E, \overset{(}{Z}_E]^T$. The input-output relations (for higher-order kinematics) between the RCM multidual parameters $\overset{(}{\psi}, \overset{(}{\theta}, \overset{(}{l}_{ins}$ and the active joints multidual parameters $\overset{(}{\rho}_i$, $i = \overline{1,3}$ result from $\overset{(}{\mathbf{V}}_1 - \overset{(}{\mathbf{V}}_2 = \mathbf{0}$; the input-output relations can be solved for $\overset{(}{\psi}, \overset{(}{\theta}, \overset{(}{l}_{ins}$ (forward kinematics) and for $\overset{(}{\rho}_i$, $i = \overline{1,3}$ (inverse kinematics).

*Algorithm 6* (*Inverse Kinematics – Multidual Homogeneous Matrix*) show how the obtain the inverse kinematic models.

*Algorithm 6 (Inverse Kinematics – Multidual Homogeneous Matrix Approach)*

---

**0. Inputs**: instrument tip parameters: $\mathbf{X}_E = [X_E, Y_E, Z_E]^T$, $\overset{(}{\mathbf{X}}_E = \mathbf{X}_E + \varepsilon \dot{\mathbf{X}}_E + \frac{\varepsilon^2}{2} \ddot{\mathbf{X}}_E + \frac{\varepsilon^3}{6} \dddot{\mathbf{X}}_E$

**1. Compute**:
$$\overset{(}{\mathbf{V}}_1 = \overset{(}{\mathbf{P}}_1 - \overset{(}{\mathbf{R}}_{RCM} \mathbf{R}_{RCM}^T \mathbf{P}_1,$$
$$\overset{(}{\mathbf{V}}_2 = \overset{(}{\mathbf{P}}_2 - \overset{(}{\mathbf{R}}_{RCM} \mathbf{R}_{RCM}^T \mathbf{P}_2,$$
$$\overset{(}{\mathbf{V}}_1 = \overset{(}{\mathbf{V}}_2.$$

**2. Solve**: $\overset{(}{\mathbf{V}}_1 - \overset{(}{\mathbf{V}}_2 = \mathbf{0}$ for $\mathbf{Q}_\rho = [\rho_1, \rho_2, \rho_3]^T$, $\overset{(}{\mathbf{Q}}_\rho$

**3. Set** $\varepsilon^k \neq 0$, $\varepsilon^n = 0$, $n > 3$, $k = \overline{1,3}$, and

**Compute**: $\dot{\mathbf{Q}}_\rho$ (evaluate $\overset{(}{\mathbf{Q}}_\rho$ for $[\varepsilon^0 = 1, \varepsilon^1 = 0, \varepsilon^2 = 0]$)

$\ddot{\mathbf{Q}}_\rho$ (evaluate $\overset{(}{\mathbf{Q}}_\rho$ for $[\varepsilon^0 = 0, \varepsilon^1 = 1, \varepsilon^2 = 0]$)

$\dddot{\mathbf{Q}}_\rho$ (evaluate $\overset{(}{\mathbf{Q}}_\rho$ for $[\varepsilon^0 = 0, \varepsilon^1 = 0, \varepsilon^2 = 1]$)

**4. Outputs**: $\mathbf{Q}_\rho = [\rho_1, \rho_2, \rho_3]^T$, $\dot{\mathbf{Q}}_\rho$, $\ddot{\mathbf{Q}}_\rho$, $\dddot{\mathbf{Q}}_\rho$

---

### 4.4. *Kinematic modeling of the hybrid parallel robot (dual quaternion approach)*

Consider the righthand side of Eq. (27) which maps into a dual quaternion $\underline{\mathbf{Q}}_P$ as (using Eqs. A24, A25 – Appendix A):

$$\begin{bmatrix} \mathbf{R}_Z(\psi) \cdot \mathbf{R}_Y(\theta) & \mathbf{R}_Z(\psi) \cdot \mathbf{R}_Y(\theta) \cdot \mathbf{d}_{L0} \\ \mathbf{0} & 1 \end{bmatrix} \to \underline{\mathbf{Q}}_P : \begin{bmatrix} x_{0P} \\ x_{1P} \\ x_{2P} \\ x_{3P} \\ y_{0P} \\ y_{1P} \\ y_{2P} \\ y_{3P} \end{bmatrix} = h \begin{bmatrix} 2 \\ -2u_1 u_2 \\ 2u_2 \\ 2u_1 \\ (l - l_{ins}) u_2 u_1 \\ (l - l_{ins}) \\ (l - l_{ins}) u_1 \\ (l - l_{ins}) u_2 \end{bmatrix} \tag{36}$$

where $u_1$, $u_2$ are the tangent of the half angles of $\psi$, $\theta$, and $h$ is a factor to ensure the quaternion normalization condition $x_0^2 + x_1^2 + x_2^2 + x_3^2 = 1$. The dual quaternion transformation that changes the perspective from **P** to **E** is given by $\underline{Q}_E = \underline{Q}_P \underline{Q}_L$, and the change from **E** to **P** is given by $\underline{Q}_P = \underline{Q}_E \underline{Q}_L^*$ where $\underline{Q}_L^*$ is the conjugate of $\underline{Q}_L$. The $\underline{Q}_E$ and $\underline{Q}_L$ dual quaternions read:

$$\underline{Q}_E : \begin{bmatrix} x_{0E} \\ x_{1E} \\ x_{2E} \\ x_{3E} \\ y_{0E} \\ y_{1E} \\ y_{2E} \\ y_{3E} \end{bmatrix} = h \begin{bmatrix} 2 \\ -2u_1 u_2 \\ 2u_2 \\ 2u_1 \\ l_{ins} u_2 u_1 \\ l_{ins} \\ l_{ins} u_1 \\ -l_{ins} u_2 \end{bmatrix}, \underline{Q}_L : \begin{bmatrix} x_{0P} \\ x_{1P} \\ x_{2P} \\ x_{3P} \\ y_{0P} \\ y_{1P} \\ y_{2P} \\ y_{3P} \end{bmatrix} = \begin{bmatrix} 1 \\ 0 \\ 0 \\ 0 \\ 0 \\ -\frac{1}{2}(l - l_{ins}) \\ 0 \\ 0 \end{bmatrix}. \tag{37}$$

A second dual quaternion $\underline{Q}_{P2}$ is defined using the substitution of $\mathbf{P}_2$ into $\mathbf{M}_{P1}$, namely:

$$\begin{bmatrix} \mathbf{R}_Z(\psi) \cdot \mathbf{R}_Y(\theta) & \mathbf{d}_{\rho 1} \cdot \mathbf{R}_Y(\rho_3) \cdot \mathbf{d}_{Q2} \\ 0 & 1 \end{bmatrix} \rightarrow \underline{Q}_{P2} : \begin{bmatrix} x_{0P2} \\ x_{1P2} \\ x_{2P2} \\ x_{3P2} \\ y_{0P2} \\ y_{1P2} \\ y_{2P2} \\ y_{3P2} \end{bmatrix} = h_1 \begin{bmatrix} 2(u_3^2 + 1) \\ -2(u_3^2 + 1)u_1 u_2 \\ 2(u_3^2 + 1)u_2 \\ 2(u_3^2 + 1)u_1 \\ \left((u_3^2 l_0 - 2u_3 \rho_2 + l_0)u_2 - \rho_2 u_3^2 + \rho_2\right)u_1 + \rho_1 u_2 (u_3^2 + 1) \\ (-\rho_1 u_1 - \rho_2 u_2 + l_0)u_3^2 - 2u_3 \rho_2 - \rho_1 u_1 + \rho_2 u_2 + l_0 \\ \left((-\rho_2 u_2 - l_0)u_3^2 + 2u_3 \rho_2 + \rho_2 u_2 - l_0\right)u_1 - \rho_1 u_3^2 - \rho_1 \\ \left((-\rho_1 u_1 + l_0)u_3^2 - 2u_3 \rho_2 - \rho_1 u_1 + l_0\right)u_2 + \rho_2 u_3^2 - \rho_2 \end{bmatrix} \tag{38}$$

where $u_3$ is the tangent of the half angle of $\rho_3$. Assume that $\underline{Q}_{P2}$ and $\underline{Q}_P$ represent the same physical displacement in SE(3), and equate them to eliminate the $x_{iPj}$, $y_{iPj}$, $i = \overline{0,3}$, $j = \overline{1,2}$ parameters. Consequently, the eight remaining mathematical relations are input-output equations that relate the RCM parameters $u_1$, $u_2$, $l_{ins}$ with the active joints parameters $\rho_1$, $\rho_2$, $u_3$. A possible way to solve these equations for either $u_1$, $u_2$, $l_{ins}$ or $\rho_1$, $\rho_2$, $u_3$ is to first eliminate the $h$, $h_1$ homogenizing parameters. This is achieved by computing a lexdeg ordering Groebner base [39] $G_0$ with [$h$, $h_1$] as the list of eliminated variables and [$\psi$, $\theta$, $l_{ins}$, $\rho_1$, $\rho_2$, $u_3$] the unknown variables. The base $G_0$ contains 24 polynomials independent of $h$ and $h_1$. Two other Groebner bases are computed thereafter from $G_0$ using a plex ordering: $G_{FK}$ using the [$u_1$, $u_2$, $l_{ins}$] ordering, which yields three polynomials (total degree being 8); $G_{IK}$ using the [$\rho_1$, $\rho_2$, $u_3$] ordering, which yields three polynomials (total degree being 4). In general form the solutions for solving $G_{IK}$ are:

$$\begin{cases} \rho_1 = f_1(u_1, u_2, l_{ins}) \\ \rho_1 = f_2(u_1, u_2, l_{ins}) \\ \rho_1 = f_3(u_1, u_2, l_{ins}) \end{cases} \tag{39}$$

where $f_1$, $f_2$, $f_3$ are analytic functions of the RCM parameters. Appendix C4 shows numerical examples regarding the mapping between $u_1$, $u_2$, $l_{ins}$ and $\rho_1$, $\rho_2$, $u_3$ parameters using the dual quaternion approach.

*4.4.1.    Higher order kinematics using dual quaternion differentiation*

The $\underline{Q}_P$ and $\underline{Q}_{P2}$ dual quaternions represent the same physical displacement. By differentiation the following relation is obtained:

$$\overset{\text{\tiny \&\&\&}}{\mathbf{q}}_{RCM} + \varepsilon_0 \tfrac{1}{2}\left(\overset{\text{\tiny \&\&\&}}{\mathbf{d}}_1 \mathbf{q}_{RCM} + \mathbf{d}_1 \overset{\text{\tiny \&\&\&}}{\mathbf{q}}_{RCM}\right) = \overset{\text{\tiny \&\&\&}}{\mathbf{q}}_{RCM} + \varepsilon_0 \tfrac{1}{2}\left(\overset{\text{\tiny \&\&\&}}{\mathbf{d}}_2 \mathbf{q}_{RCM} + \mathbf{d}_2 \overset{\text{\tiny \&\&\&}}{\mathbf{q}}_{RCM}\right),$$

$$\mathbf{q}_{RCM} = \begin{bmatrix} 2 \\ -2u_1 u_2 \\ 2u_2 \\ 2u_1 \end{bmatrix}, \mathbf{d}_1 = \begin{bmatrix} 0 \\ l_{ins} - l \\ 0 \\ 0 \end{bmatrix}, \mathbf{d}_2 = \begin{bmatrix} 0 \\ -l_0 + \cos(\rho_3) \\ \rho_1 \\ \rho_2 \sin(\rho_3) \end{bmatrix}. \tag{40}$$

and using Eq. (14) all the higher derivatives (assuming they exist) can be computed. Eq. (40) and all subsequent derivatives can be solved for the higher order kinematics of the active joints parameters. *Algorithm 7*(*Inverse Kinematics – Dual Quaternion Differentiation*) show how the obtain the inverse kinematic models.

*Algorithm 7 (Inverse Kinematics – Dual Quaternion Differentiation)*

---

0. **Inputs**: instrument tip parameters: $\underline{\mathbf{Q}}_E, \overset{\text{\tiny \&}}{\underline{\mathbf{Q}}}_E, \overset{\text{\tiny \&\&}}{\underline{\mathbf{Q}}}_E, \overset{\text{\tiny \&\&\&}}{\underline{\mathbf{Q}}}_E$

1. **Compute**: $\underline{\mathbf{Q}}_P = \underline{\mathbf{Q}}_E \underline{\mathbf{Q}}_L^*$

2. **Compute**: $\underline{\mathbf{Q}}_P^{(n)}$, and $\underline{\mathbf{Q}}_{P2}^{(n)}$ using Eq. (14) applied on Eqs. (36), (38)

3. **Solve**: $\underline{\mathbf{Q}}_P^{(n)} - \underline{\mathbf{Q}}_{P2}^{(n)} = \mathbf{0}$, for $\mathbf{Q}_\rho^{(n)} = [\rho_1^{(n)}, \rho_2^{(n)}, u_3^{(n)}]^T$, $n = \overline{1,3}$

4. **Outputs**: $\mathbf{Q}_\rho^{(n)} = [\rho_1^{(n)}, \rho_2^{(n)}, u_3^{(n)}]^T$, $n = \overline{0,3}$

---

To compute the inputs of Algorithm 7 Eq. (37) is used.

#### 4.4.2. Higher order kinematics using the HMD quaternion

To achieve the higher order kinematic using the HMD quaternions, a similar approach as with homogeneous matrices is taken. Based on the general formulation in Eq. (14) and the dual quaternion from Eq. (28), the HMD quaternion that represents the higher order kinematics of point P with respect to the RCM is shown in Eq. (31.a). Furthermore, the HMD quaternion that represents the higher order kinematics of point P with respect to the active joint parameters is shown in Eq. (31.b).

$$\overset{(}{\underline{\mathbf{Q}}}_P = \left(1 + \tfrac{1}{2}\varepsilon_0 \mathbf{d}_1\right) \overset{(}{\mathbf{q}}_{RCM}$$

$$\mathbf{q}_{RCM} = \begin{bmatrix} 2 \\ -2u_1 u_2 \\ 2u_2 \\ 2u_1 \end{bmatrix}, \mathbf{d}_1 = \begin{bmatrix} 0 \\ l_{ins} - l \\ 0 \\ 0 \end{bmatrix} \tag{41.a}$$

$$\overset{(}{\underline{\mathbf{Q}}}_{P2} = \left(1 + \tfrac{1}{2}\varepsilon_0 \mathbf{d}_2\right) \overset{(}{\mathbf{q}}_{RCM}$$

$$\mathbf{q}_{RCM} = \begin{bmatrix} 2 \\ -2u_1 u_2 \\ 2u_2 \\ 2u_1 \end{bmatrix}, \mathbf{d}_2 = \begin{bmatrix} 0 \\ -l_0 + \cos(\rho_3) \\ \rho_1 \\ \rho_2 \sin(\rho_3) \end{bmatrix}. \tag{41.b}$$

Assuming that $\underline{\mathbf{Q}}_{P2}$ and $\underline{\mathbf{Q}}_P$ represent the same Euclidean displacement, it follows that $\overset{(}{\underline{\mathbf{Q}}}_P - \overset{(}{\underline{\mathbf{Q}}}_{P2} = \mathbf{0}$ yields input-output relations that can be solved for the RCM parameters or for the active joint parameters. *Algorithm 8*(*Inverse Kinematics – HMD*) show how the obtain the inverse kinematic models.

*Algorithm 8 (Inverse - Kinematics HMD)*

---

0. **Inputs**: instrument tip parameters: $\underline{\mathbf{Q}}_E, \overset{(}{\underline{\mathbf{Q}}}_E$

1. **Compute**: $\underline{\mathbf{Q}}_P = \underline{\mathbf{Q}}_E \underline{\mathbf{Q}}_L^*$

2. **Compute**: $\overset{(}{\underline{\mathbf{Q}}}_P$, and $\overset{(}{\underline{\mathbf{Q}}}_{P2}$ using Eqs. (41.a), (41.b)

3. Solve: $\overset{(}{\underline{Q}}_p - \overset{(}{\underline{Q}}_{p2} = \mathbf{0}$, for $\overset{(}{\underline{Q}}_p = [\overset{(}{\rho}_1, \overset{(}{\rho}_2, \overset{(}{u}_3]^T$,

4. Set $\varepsilon^k \neq 0$, $\varepsilon^n = 0$, $n > 3$, $k = \overline{1,3}$, and

   **Compute**: $\overset{\&(}{Q}_p$ (evaluate $\overset{\&(}{Q}_p$ for $[\varepsilon^0 = 1, \varepsilon^1 = 0, \varepsilon^2 = 0]$)

   $\overset{\&\&(}{Q}_p$ (evaluate $\overset{\&(}{Q}_p$ for $[\varepsilon^0 = 0, \varepsilon^1 = 1, \varepsilon^2 = 0]$)

   $\overset{\&\&\&(}{Q}_p$ (evaluate $\overset{\&(}{Q}_p$ for $[\varepsilon^0 = 0, \varepsilon^1 = 0, \varepsilon^2 = 1]$)

4. **Outputs**: $\mathbf{Q}_p^{(n)} = [\rho_1^{(n)}, \rho_2^{(n)}, u_3^{(n)}]^T$, $n = \overline{0,3}$

To compute the inputs of Algorithm 7 Eq. (37) is used.

## 5. Numerical results and simulations

### 5.1. Numerical evaluation and comparison methodology

All kinematic models were computed in Maple and exported in MATLAB. Two expression optimizations were applied for all symbolic calculations: i) simplify at each computation step; ii) optimize command for the last step in the computation (where the general expression for an output parameter was computed). The main targeted comparison of the kinematic algorithms was between the algorithms implemented using differentiation and the ones implemented using multidual algebra (comparing each formalism individually). The validity of all the inverse kinematic algorithms was tested with the forward kinematics (although these algorithms were not presented in the paper) by testing that an arbitrary input trajectory in the WS for inverse kinematics yields a output trajectory which in turn was provided into the forward kinematics and produce the original WS trajectory.

The following quantitative measuring parameters are defined together with the numerical evaluation tests:

1. **Accuracy** of the inverse kinematic algorithms with respect to a defined trajectory. To measure this parameter, the following time dependent multiparametric functions were defined:

   a. $\gamma_1(t) = [\rho_i(t), \overset{\&}{\rho}_i(t), \overset{\&\&}{\rho}_i(t), \overset{\&\&\&}{\rho}_i(t)]$, $i = \overline{1,3}$ defined as input for Algorithms 1, and $\gamma_2(t) = [\overset{(}{\rho}_i(t)]$, $i = \overline{1,3}$ (derived based on $\gamma_1(t)$ such that they represent the same trajectory). The outputs for Algorithms 1, 2 are defined as $\mathbf{Q}_1(t) = [q_i(t), \overset{\&}{q}_i(t), \overset{\&\&}{q}_i(t), \overset{\&\&\&}{q}_i(t)]$, $i = \overline{1,3}$ and $\mathbf{Q}_2(t) = [q_i(t), \overset{\&}{q}_i(t), \overset{\&\&}{q}_i(t), \overset{\&\&\&}{q}_i(t)]$, $i = \overline{1,3}$, respectively.

   b. $\gamma_2(t) = [\mathbf{E}(t), \overset{\&}{\mathbf{E}}(t), \overset{\&\&}{\mathbf{E}}(t), \overset{\&\&\&}{\mathbf{E}}(t)]$ defined as input for Algorithms 3, 4, 5, 6. The outputs for Algorithms 3, 4, 5, 6 are defined as $\delta_j(t) = [\rho_i(t), \overset{\&}{\rho}_i(t), \overset{\&\&}{\rho}_i(t), \overset{\&\&\&}{\rho}_i(t)]$, $i = \overline{1,3}$, $j = \overline{1,4}$.

   c. $\gamma_3(t) = [\mathbf{Q}_E(t), \overset{\&}{\mathbf{Q}}_E(t), \overset{\&\&}{\mathbf{Q}}_E(t), \overset{\&\&\&}{\mathbf{Q}}_E(t)]$ defined as input for Algorithms 7, 8. The outputs are defined as $\delta_j(t) = [\rho_i(t), \overset{\&}{\rho}_i(t), \overset{\&\&}{\rho}_i(t), \overset{\&\&\&}{\rho}_i(t)]$, $i = \overline{1,3}$, $j = \overline{5,6}$.

   The residuals of $\mathbf{Q}_1(t) - \mathbf{Q}_2(t)$, $\delta_1(t) - \delta_2(t)$, $\delta_3(t) - \delta_4(t)$, $\delta_5(t) - \delta_6(t)$, are studied to determine if there are statistically significant differences between multidualization and differentiation for the kinematic algorithms.

2. Estimated **Execution Time** of the algorithm. This parameter was studied for pairs of algorithms: Algorithm 1 was compared with 2, 3 with 4, 5 with 6 and, 7 with 8. The tests were defined as looping 1000 experimental trials, where for each experimental trial, 1000 discrete sample vectors (each vector containing displacement, velocity, acceleration, jerk) from a WS input trajectory was given. Each experimental trial execution time was estimated with timeit() function (MATLAB).

**Note:** All the numerical tests were performed on a PC with: 12th Gen Intel(R) Core(TM) i9-12900K 3.20 GHz, 32.0 GB RAM, running Windows 10 Pro.

### 5.2. Numerical results for the comparison

In the **Accuracy** test the residuals of $\mathbf{Q}_1(t) - \mathbf{Q}_2(t)$, $\delta_1(t) - \delta_2(t)$, $\delta_3(t) - \delta_4(t)$, $\delta_5(t) - \delta_6(t)$, were studied with the following results:

1. For $\mathbf{Q}_1(t) - \mathbf{Q}_2(t)$, $\boldsymbol{\delta}_1(t) - \boldsymbol{\delta}_2(t)$, $\boldsymbol{\delta}_3(t) - \boldsymbol{\delta}_4(t)$ the root mean square (RMS) between the output results was in the order of 1.0e–16 for the acceleration and jerk fields, and 0 for the velocity and displacement fields.
2. For $\boldsymbol{\delta}_5(t) - \boldsymbol{\delta}_6(t)$ the RMS between the outputs was in the order of 1.0e–12 for the acceleration and jerk fields, 1.0e–13 for the velocity field, and 0 for displacement.
3. For all the tested residuals, the autocorrelation test showed no correlation indicating the small differences are due to random noise.

For the **Execution Time** estimation, Figure 4.a shows the execution times for each experimental trial benchmarking Algorithms 1 and 2 (the algorithms that create the mapping between the $q_i$, $i = \overline{1,3}$ and $\rho_i$, $i = \overline{1,3}$ parameters). Figure 4.b shows the execution times for Algorithms 3 and 4 which are implemented with differentiating and multidualization of the kinematic Jacobian. Figure 4.c shows the execution times for Algorithms 5 and 6 which are implemented with differentiation and multidualization of the homogeneous transformation matrix. Lastly, Figure 4.d shows the execution times for Algorithms 7 and 8 which are implemented based on dual quaternion differentiation and HMD. As evident from Figure 4, while both classical differentiation and multidualization approaches have efficient execution times, the multidual approach is statistically better in execution time. Furthermore, in most cases the multidual approach seems to be more stable as it shows fewer spikes in execution time.

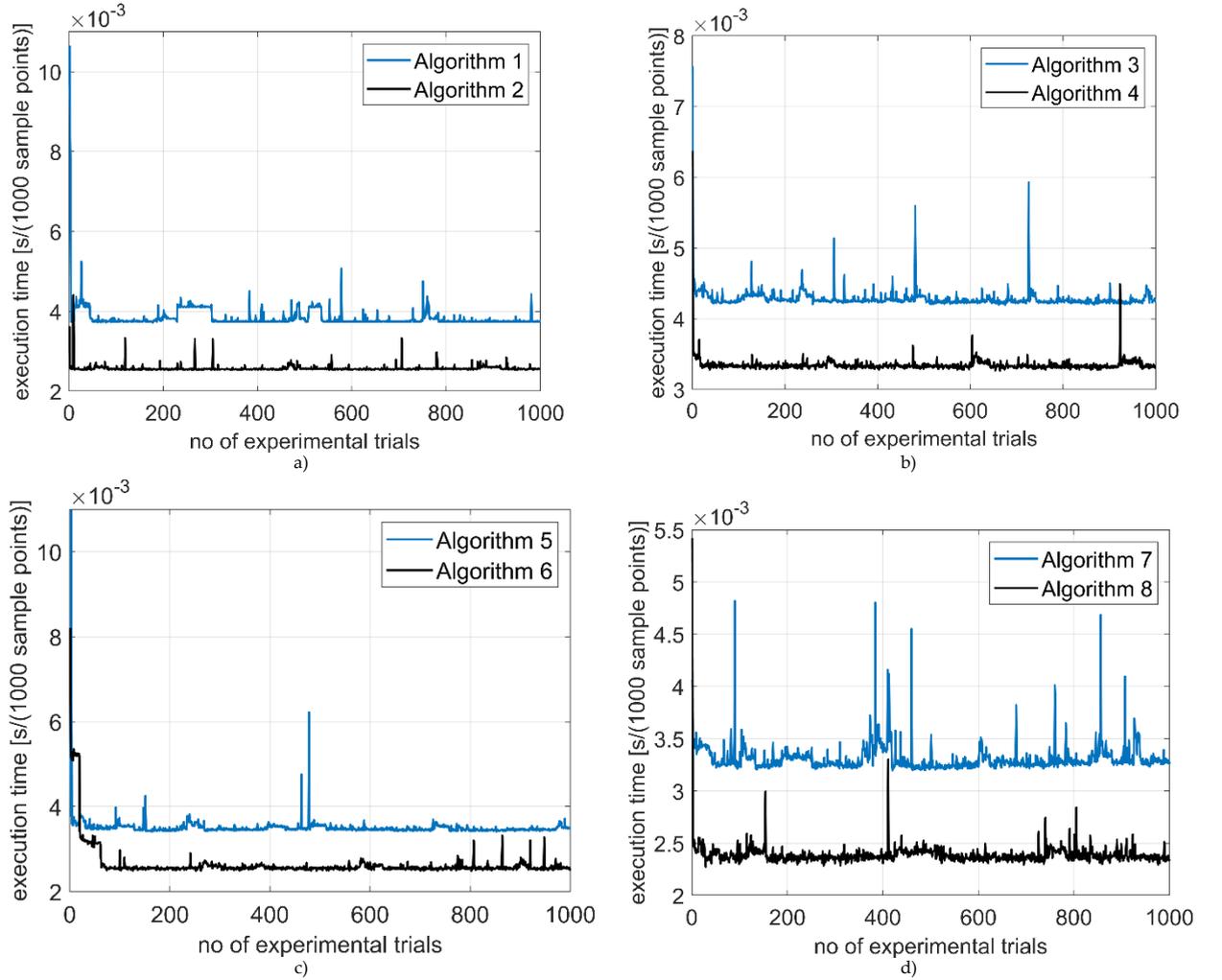

**Figure 4**. Execution times for evaluating kinematic algorithms implemented based on classical differentiation and multidual algebra: a) Algorithm 1 compared with Algorithm 2; b) Algorithm 3 compared with Algorithm 4; c) Algorithm 5 compared with Algorithm 6; d) Algorithm 7 compared with Algorithm 8.

## 5.3. Numerical simulations

To assess the efficiency of the proposed algorithms, a simulation of the surgical instrument re-positioning and re-orientation has been performed, using the proposed multidual algebra algorithms in MATLAB and the CAD model developed in Siemens NX, using the following input trajectory:

$$\begin{cases} X_{E\_start} = 5.88\,mm \\ Y_{E\_start} = 0.14\,mm \\ Z_{ins\_i} = -5.02\,mm \end{cases} \begin{cases} X_{E\_finaal} = 82.59\,mm \\ Y_{E\_start} = 14.56\,mm \\ Z_{ins\_i} = -54.46\,mm \end{cases} or \begin{cases} \psi_{start} = 1.40^o \\ \theta_{start} = 40.44^o \\ L_{ins\_initial} = 7.73\,mm \end{cases} \begin{cases} \psi_{final} = 10^o \\ \theta_f = 33^o \\ L_{ins\_final} = 100\,mm \end{cases} \quad (42)$$

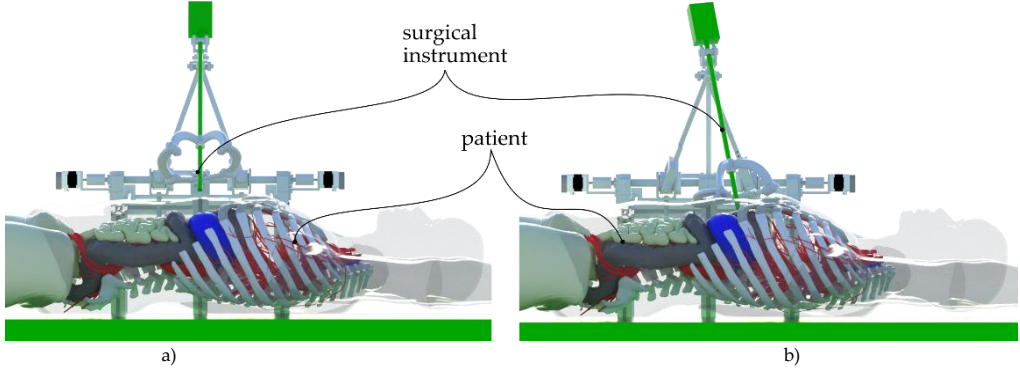

**Figure 5.** CAD model of the hybrid parallel robot during the simulation: a) initial pose; b) final pose.

The motion law for the end-effector, point **E**, is given by: $a_{max} = 4\,mm/s^2$; $j_{max} = 2\,mm/s^3$. The CAD model of the robot in the initial and final pose of the simulation is illustrated in Figure 5. Since the simulation was achieved for the kinematics of the proposed modular parallel robot, no data regarding the material density, friction, etc. were required.

Figure 6 presents the time history diagram of the displacements, velocities, accelerations and jerks of the input data into the simulation. A trapezoidal profile for the acceleration of the surgical instrument tip $\dddot{X}_E, \dddot{Y}_E, \dddot{Z}_E$ has been used, while the derived trajectory of the end-effector is constrained by the RCM position (also the origin of the Cartesian frame of the robot). The time history diagram of the active joints' coordinates $q_1, q_2, q_3$ of the parallel robot for displacements, velocities, accelerations and jerks is presented in Figure 7.

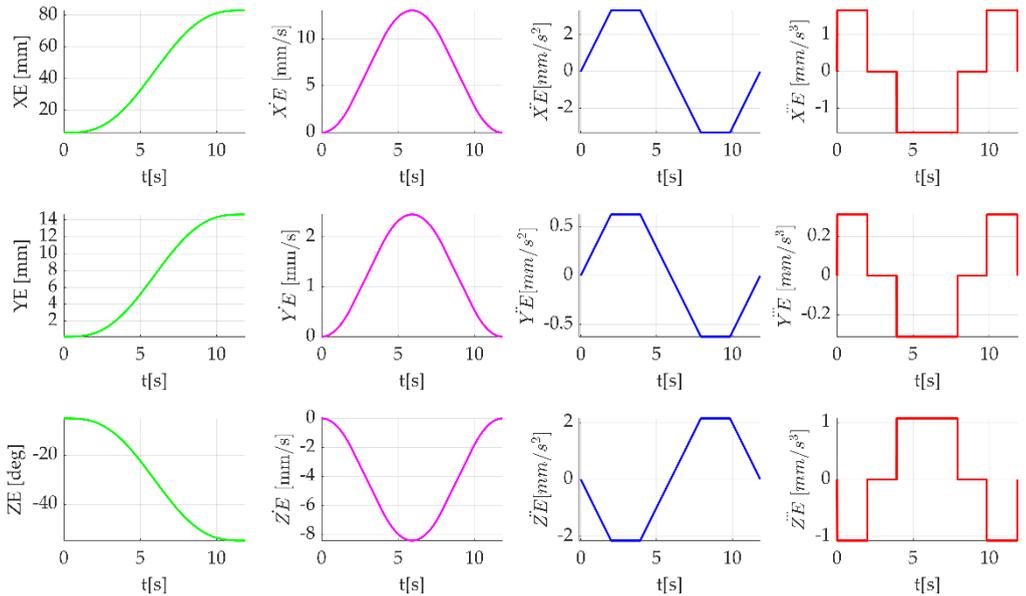

Figure 6. The time history diagram for the end-effector: a) displacements; b) velocities; c) accelerations; d) jerks.

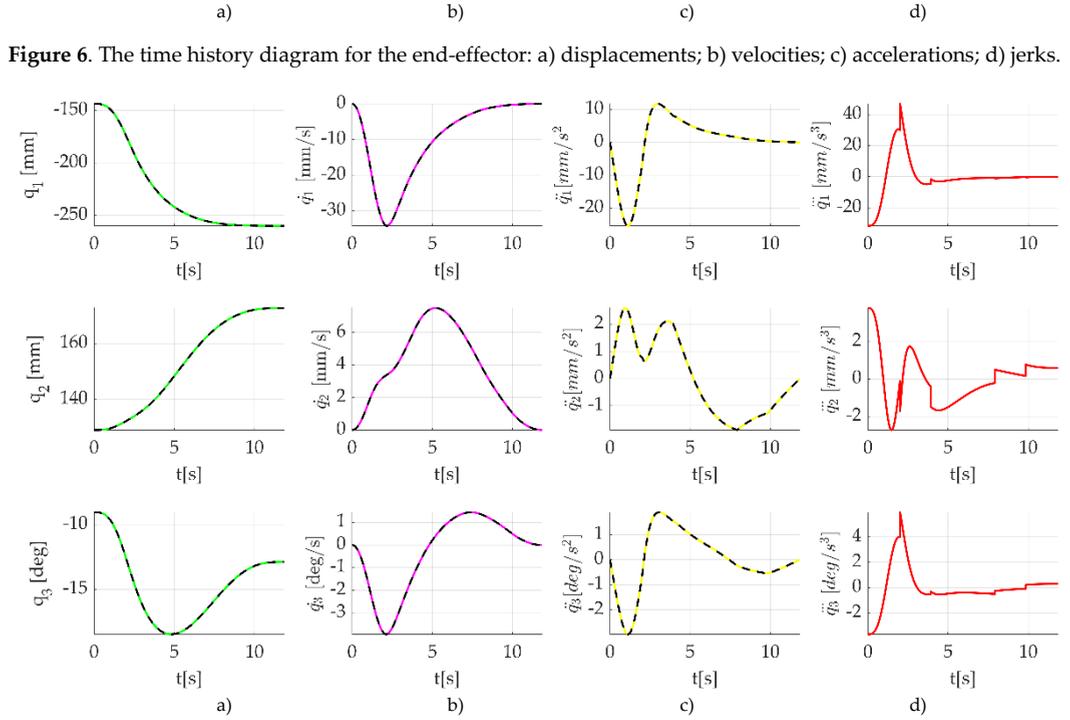

Figure 7. The time history diagram for the parallel robot active joints: a) displacements; b) velocities; c) accelerations; d) jerks.

Using the Siemens NX CAD model the displacements, velocities and accelerations of the active joints have also been generated and the results superposed in Figure 7 (MATLAB in continuous lines and NX using dashed black lines). The superposition of the computed curves using MATLAB and the ones using Siemens NX show a very good correlation (for example, for displacements it yields $RMSE_{q1} = 0.15, RMSE_{q2} = 0.12, RMSE_{q3} = 0.13$ ), which validates the determined kinematic model. The input trajectory (Figure 6) was numerically generated; the jerk curves profiles are defined as a step signal where the transition points are discontinued (they are defined using two points at the same instance of time). This in turn, induces the spike behavior in the jerk curves in Figure 7 (which are computed based on symbolic equations).

## 6. Discussions

Based on the numerical analysis presented in Section 5 there are a few important points to highlight for the higher-order kinematics of the hybrid parallel robot for pancreatic surgery:

1. There is a statistically significant difference in execution times between the differentiation and multidualization approach. In addition, Figure 4 shows that the multidual approach has fewer time spikes which can generate stability. The limitation of the presented results is that the benchmark was performed on a standard PC, and the differences in execution times are negligible for real time control. However, due to the test consistency, the claim that multidual implementation will have an even greater effect on robot control hardware is justified.
2. Both the differentiating and multidual implementations are comparable in terms of accuracy; the RMS values when comparing algorithms were not statistically significant and the autocorrelation tests indicated that the small differences between the algorithms outputs are due to random (numerical) noise.
3. The better execution time for the multidual approaches can be attributed to the sequence of evaluation of the symbolic equations for the higher-order kinematics (e.g., Algorithm 2 – step 2), where the input kinematic parameters (displacement, velocity, acceleration, jerk) are substituted first (and only once), and then the values for the output kinematic parameters are computed by setting the appropriate values for $\varepsilon^i, i = \overline{0,n}$ (e.g., setting $\varepsilon^1 = 1, \varepsilon^i = 0, i = \{0, \overline{2,n}\}$ computes velocities, $\varepsilon^2 = 1, \varepsilon^i = 0, i = \{0,1,\overline{3,n}\}$ computes accelerations). This approach is fundamentally different than the step-by-step substitution required in the classical approaches where the kinematic parameters are substituted at each computation step.

Despite having a single case study for the proposed kinematic algorithms, the consistency between the results leads to the hypothesis that multidual approaches may generally perform better than differentiation approaches, even in symbolic expressions.

When discussing symbolic calculus, the advantages of multidual algebra over the other formalisms of higher order kinematics are qualitative. First, multidual algebra represents a unifying framework for higher order kinematics, i.e., it's used (at least in this work) as a mathematical tool to compute the higher order kinematics regardless of the formalism adopted for the rigid body displacement. Secondly, multidual algebra computes simultaneously all the higher order kinematics up to a degree n. Thirdly, multidual algebra is compact. With access to mathematical tools such as multidualization and HMD quaternions can simplify the computation of higher order kinematics, and although it was not shown in this paper, it is safe to assume that these tensor methods are easily implemented in vector programming paradigms (e.g., MATLAB) as straight numerical algorithms. Furthermore, higher order kinematics facilitate the dynamic modelling of parallel robots [40], where in the particular case of surgery, is useful in trajectory planning to generate dynamic models that anticipate and handle abrupt changes in velocity and acceleration and the optimization of actuator forces.

## 7. Conclusions

The paper presents the higher-order kinematics of a parallel robot for minimally invasive surgical applications. It features a modular architecture consisting of a spherical parallel mechanism with an RCM and a 4-DOF parallel robot which actuates the surgical instrument. The paper takes advantage of the modular architecture of the parallel robot and breaks the kinematic analysis into several stages: the study of the RCM, namely the points connected to the surgical instrument ($E, RCM, P$), the serial parametrization of the parallel robot (using the $\rho_i, i=\overline{1,3}$ coordinates) and the actual active coordinates of the parallel robot $q_i, i=\overline{1,3}$. The fourth active coordinate has been disregarded in the kinematic analysis, since the $\varphi$ angle orientation of the surgical instrument is directly connected to this coordinate ($\varphi = q_4$). This has some advantages: the RCM kinematic analysis can be used in the case of any mechanism with an active RCM, the differences being obvious after the determination of the $P$ point kinematics, and it makes it easier to explain and compute the higher-order acceleration fields.

The inverse kinematics of higher order have been developed using three different approaches: the classical vector method which leads to the velocities, accelerations, etc. using the Jacobian, and its time successively obtained derivatives, the multidual algebra, which has been used in three different scenarios: the multidualization of the Jacobian, and the multidualization of the homogenous transformation matrix, and the hyper multidual quaternion approach, which uses the same multidual algebra for the parametrization of the higher order kinematics using dual quaternions.

The results showed that the implementation of kinematic functions using multidual algebra has statistically significant better execution times, while remaining as accurate as the differentiation methods.

Further work is required to implement and test the kinematic functions of robot control hardware, in order to validate the approach.

**Acknowledgements**

This research was supported by the project New smart and adaptive robotics solutions for personalized minimally invasive surgery in cancer treatment - ATHENA, funded by European Union – NextGenerationEU and Romanian Government, under National Recovery and Resilience Plan for Romania, contract no. 760072/23.05.2023, code CF 116/15.11.2022, through the Romanian Ministry of Research, Innovation and Digitalization, within Component 9, investment I8.

**APPENDIX A. Multidual algebra background**

*A.1 Multidual numbers and functions of multidual arguments*

Considering a real number $x, x_1 \in \mathbb{R}$, a dual number is defined as, [32,34-36]:

$$\underline{x} = x + \varepsilon_0 \cdot x_1, \varepsilon_0 \neq 0, \varepsilon_0^2 = 0. \tag{A1}$$

The set of multidual numbers is further defined as:

$$\underline{x} = x + \varepsilon \cdot x_1 + \varepsilon^2 \cdot x_2 + \mathrm{K} + \varepsilon^n \cdot x_n, \ \varepsilon \neq 0, \ \varepsilon^{n+1} = 0, \ n \in \mathbb{N}^*. \tag{A2}$$

where $x, x_i \in \mathbb{R}, i = \overline{1,n}$, $x = \mathrm{Re}(\underline{x})$ and $\sum_{k=1}^{n} x_k \varepsilon^k = \mathrm{MD}(\underline{x}) = \underline{x} - x$, $\mathrm{Re}(\underline{x})$ being the real part of $\underline{x}$ and $\mathrm{MD}(\underline{x})$ the multidual part of $\underline{x}$. Considering an n-differentiable real function: $h : \mathfrak{I} \subset \mathbb{R} \to \mathbb{R}$ with $h = h(x)$, the multidual function depending on the multidual variable $\underline{x}$ is:

$$h(\underline{x}) = h(x) + \sum_{k=1}^{n} \frac{\Delta(\underline{x})^k}{k!} h^{(k)}(x), \tag{A3}$$

where $\Delta(\underline{x}) = \underline{x} - x = \sum_{k=1}^{n} x_k \varepsilon^k, \varepsilon \neq 0, \varepsilon^{n+1} = 0$.

If $h : \mathfrak{I} \subset \mathbb{R} \to \mathbb{R}, h = h(t)$ is time-dependent, n-differentiable real function, the multidual associated function, named multidual differential transform, is:

$$\breve{h} = h + \varepsilon \dot{h} + \frac{\varepsilon^2}{2!} \ddot{h} + \mathrm{K} + \frac{\varepsilon^n}{n!} h^{(n)} = e^{\varepsilon D} h, \ n \in \mathbb{N}^*, \tag{A4}$$

where $e^{\varepsilon D} = 1 + \varepsilon D + \mathrm{K} + \frac{\varepsilon^n}{n!} D^n$ and $D = \frac{d}{dt}$ as the time derivative operator.

## 7.1. A.2 Properties of functions of multidual differential transform

Let $f$ and $g$ be functions of a multidual transform (A4), The following are true:

$$\begin{aligned} \widetilde{f+g} &= \breve{f} + \breve{g}, \\ \widetilde{fg} &= \breve{f}\breve{g}, \\ \widetilde{\lambda f} &= \lambda \breve{f}, \forall \lambda \in \mathbb{R}, \\ \widetilde{f(\alpha)} &= f(\breve{\alpha}), \alpha \in C^n(I) \\ \breve{\breve{f}} &= \breve{f} \end{aligned} \tag{A5}$$

$f, g, I \subset \mathbb{R} \to \mathbb{R}$ are n-times differentiable functions.

## A.3 The multidual kinematic Jacobian

. Considering a lower-pair chain described by the following kinematic mapping:

$$F = h_j(q) = \exp(\mathbf{X}_1 q_1) \cdot \exp(\mathbf{X}_2 q_2) \cdot \mathrm{K} \cdot \exp(\mathbf{X}_j q_j), j \in \mathbb{N}^*, \tag{A6}$$

where $\mathbf{X}_k, k = \overline{1,j}$ is the screw coordinate vectors and $q_k, k = \overline{1,j}$ is the joint variable.

The Jacobian of the end-effector of the kinematic chain using Eq. (A6) kinematic mapping is:

$$\mathbf{J} = \left[ \mathbf{T}_1, \mathbf{T}_2, \mathrm{K}, \mathbf{T}_j \right], \tag{A7}$$

Where $\mathbf{T}_1 = \mathbf{X}_1$ and $\mathbf{T}_k = \mathrm{Ad}_{h_{k-1}} \mathbf{X}_k, k = \overline{2,j}$ is the instantaneous screw coordinate of joint $k$. The multi dualization of the Jacobian can be easily achieved as:

$$\breve{\mathbf{J}} = \left[ \breve{\mathbf{T}}_1, \breve{\mathbf{T}}_2, \mathrm{K}, \breve{\mathbf{T}}_j \right], \tag{A8}$$

in which:

$$\overset{(}{\mathbf{T}}_k = \mathbf{T}(\overset{(}{q}_1, \overset{(}{q}_2, \ldots, \overset{(}{q}_k)).$$
(A9)

From Eq. (A3) and (A4) it yields:

$$\overset{(}{\mathbf{J}} = \mathbf{J} + \varepsilon \dot{\mathbf{J}} + \frac{\varepsilon^2}{2!}\ddot{\mathbf{J}} + K + \frac{\varepsilon^n}{n!}\mathbf{J}^{(n)}.$$
(A10)

*A.4 Multidual homogeneous transformation matrices*

In the case of tensors, one can consider a homogenous matrix which describes a rigid motion on a curve in the Lie group of rigid displacements SE(3):

$$\mathbf{b} = \begin{bmatrix} \mathbf{R} & \mathbf{r} \\ \mathbf{0} & 1 \end{bmatrix}, \mathbf{R} \in SO(3), \mathbf{R} = \mathbf{R}(t), \mathbf{r} = \mathbf{r}(t).$$
(A11)

The vector field of higher accelerations is:

$$\mathbf{H}_n = \begin{bmatrix} \mathbf{K}_n & \mathbf{v}_n \\ \mathbf{0} & 1 \end{bmatrix},$$
(A12)

where the tensor $\mathbf{K}_n$ and the vector $\mathbf{v}_n$ are obtained from:

$$\mathbf{K}_n = \mathbf{R}^{(n)}\mathbf{R}^T, \quad \mathbf{v}_n = \mathbf{r}^{(n)} - \mathbf{K}_n \mathbf{r}.$$
(A13)

where $\mathbf{a}_n(\mathbf{p}) = \mathbf{v}_n + \mathbf{K}_n \mathbf{p}$ is the higher-order acceleration vector field, $\mathbf{p} \in \mathbb{R}^3$. The multidual differential transform of $\mathbf{R}$ can be obtained from:

$$\overset{(}{\mathbf{R}} = \mathbf{R} + \varepsilon \dot{\mathbf{R}} + \frac{\varepsilon^2}{2!}\ddot{\mathbf{R}} + K + \frac{\varepsilon^n}{n!}\mathbf{R}^{(n)}.$$
(A14)

Considering the following multidual matrix:

$$\overset{(}{\mathbf{b}} = \begin{bmatrix} \overset{(}{\mathbf{R}} & \overset{(}{\mathbf{r}} \\ \mathbf{0} & 1 \end{bmatrix}.$$
(A15)

and

$$\overset{(}{\mathbf{H}} = \begin{bmatrix} \overset{(}{\mathbf{R}} & \overset{(}{\mathbf{r}} \\ \mathbf{0} & 1 \end{bmatrix} \cdot \begin{bmatrix} \mathbf{R}^T & -\mathbf{R}^T \mathbf{r} \\ \mathbf{0} & 1 \end{bmatrix},$$
(A16)

leading to:

$$\overset{(}{\mathbf{H}} = \begin{bmatrix} \overset{(}{\mathbf{K}} & \overset{(}{\mathbf{v}} \\ \mathbf{0} & 1 \end{bmatrix},$$
(A17)

where:

$$\overset{(}{\mathbf{K}} = \overset{(}{\mathbf{R}}\mathbf{R}^T, \quad \overset{(}{\mathbf{v}} = \overset{(}{\mathbf{r}} - \overset{(}{\mathbf{R}}\mathbf{R}^T \mathbf{r},$$
(A18)

which leads to:

$$\overset{(}{\mathbf{K}} = \mathbf{I} + \varepsilon \mathbf{K}_1 + \frac{\varepsilon^2}{2!}\mathbf{K}_2 + K + \frac{\varepsilon^n}{n!}\mathbf{K}_n,$$

$$\overset{(}{\mathbf{v}} = \varepsilon \mathbf{v}_1 + \frac{\varepsilon^2}{2!}\mathbf{v}_2 + K + \frac{\varepsilon^n}{n!}\mathbf{v}_n.$$
(A19)

where $\overset{(}{\mathbf{a}}(\mathbf{p}) = \overset{(}{\mathbf{v}} + \overset{(}{\mathbf{K}}\mathbf{p}$ is the compact form of the higher-order acceleration field expressed using multidual algebra, $\mathbf{p} \in \mathbb{R}^3$.

## A.5 Hyper Multidual Quaternions (HMD)

Any Euclidean displacement $D \in SE(3)$ can be mapped into a point $P$ from the projective space $P^7$ through the mapping $m$ [37]:

$$m : D \to P \in P^7$$
$$Q(x_i, y_i) \to [x_0 : x_1 : x_2 : x_3 : y_0 : y_1 : y_2 : y_3]^T \neq [0:0:0:0:0:0:0:0]^T \quad \text{(A20)}$$

The $x_i : y_i, i = \overline{0,3}$ are so-called the Study parameters which are based on a dual quaternion:

$$Q = q + \varepsilon_0 \underline{q},$$
$$q = x_0 + ix_1 + jx_2 + kx_3,$$
$$\underline{q} = y_0 + iy_1 + jy_2 + ky_3, \quad \text{(A21)}$$
$$\varepsilon_0 \neq 0, \varepsilon_0^2 = 0,$$

with the following operations defined:

$$\varepsilon_0 \neq 0, \varepsilon_0^2 = 0,$$
$$i^2 = j^2 = k^2 = -1, \; i \cdot j = k, \; j \cdot k = i, \; k \cdot i = j, \; j \cdot i = -k, \; k \cdot j = -i, \; i \cdot k = -j. \quad \text{(A22)}$$

At least one $x_i \neq 0$ is required to avoid the trivial solution (which has no use in kinematics). The quaternion $q$ describes orientations whereas its dual part $\underline{q}$ encodes the translations coupled with orientations. The dual quaternion components (Study parameters) must satisfy the unit quaternion normalizing condition and the orthogonality between the quaternion $q$ and its dual part $\underline{q}$:

$$x_0^2 + x_1^2 + x_2^2 + x_3^2 = 1$$
$$x_0 y_0 + x_1 y_1 + x_2 y_2 + x_3 y_3 = 0 \quad \text{(A23)}$$

The dual quaternion $Q$ components $x_i : y_i$ $(i = \overline{0,3})$ can be computed starting from a homogenous 4x4 transformation matrix from $SE(3)$:

$$T = \begin{bmatrix} R & \rho \\ 0 & 1 \end{bmatrix}, R \in SO(3), \rho \in \mathbb{R}^3,$$

$$R = \begin{bmatrix} r_{11} & r_{12} & r_{13} \\ r_{21} & r_{22} & r_{23} \\ r_{31} & r_{32} & r_{33} \end{bmatrix}, \rho = \begin{bmatrix} p_x \\ p_y \\ p_z \end{bmatrix}. \quad \text{(A24)}$$

in four ways:

$$x_0 : x_1 : x_2 : x_3 = 1 + r_{11} + r_{22} + r_{33} : r_{32} - r_{23} : r_{13} - r_{31} : r_{21} - r_{12}, r_{32} - r_{23} : 1 + r_{11} - r_{22} - r_{33} : r_{12} - r_{21} : r_{31} + r_{13},$$
$$r_{13} - r_{31} : r_{12} + r_{21} : 1 - r_{11} + r_{22} - r_{33} : r_{23} + r_{32}, r_{21} - r_{12} : r_{31} + r_{13} : r_{23} + r_{32} : 1 - r_{11} - r_{22} + r_{33}. \quad \text{(A25)}$$

At least one ratio from Eq. (A25) is guaranteed to be a non-singular parametrization of $SE(3)$ (if the transformation matrix does not describe a rotation by $\pi$, the first ratio can be used). The $y_i$ parameters are computed using the $x_i$ $(i = \overline{0,3})$ parameters and the translational vector $\rho$ as:

$$y_0 = -\frac{1}{2}(p_z x_3 + p_y x_2 + p_x x_1), y_1 = -\frac{1}{2}(p_z x_2 - p_y x_3 + p_x x_0),$$
$$y_2 = -\frac{1}{2}(-p_z x_1 - p_y x_0 + p_x x_3), y_3 = -\frac{1}{2}(-p_z x_0 + p_y x_1 - p_x x_2). \quad \text{(A26)}$$

In the case of the dual quaternions approach, a Hyper Multidual Quaternion (HMD) can be defined as an associated pair of an HMD scalar quantity and a free HMD vector:

$$\hat{\underline{q}} = (\hat{\underline{q}}, \hat{\vec{q}}), \hat{\underline{q}} \in \mathbb{R}, \hat{\vec{q}} \in \hat{V}_3 \quad \text{(A}$$

27)

If $\underline{\hat{u}}$ is the set of units multidual quaternions and $\underline{\hat{u}}$ is the set of units HMD quaternions, the following is true:

$$\underline{\hat{q}} = \left(1 + \varepsilon_0 \frac{1}{2}\hat{r}\right)\hat{q} \tag{A28}$$

where $\hat{r} \in \hat{V}_3$ is a multidual vector and $\hat{q} \in \hat{U}$ is a multidual unit quaternion. A hyper multidual number $\underline{\hat{\alpha}}$ and a unit hyper multidual vector $\underline{\hat{u}}$ exists, so that:

$$\underline{\hat{q}} = \cos\frac{\underline{\hat{\alpha}}}{2} + \underline{\hat{u}}\sin\frac{\underline{\hat{\alpha}}}{2} = \exp\left(\frac{\underline{\hat{\alpha}}}{2}\underline{\hat{u}}\right) \tag{A29}$$

where $\underline{\hat{\alpha}}$ and $\underline{\hat{u}}$ are the natural HMD invariants of the unit HMD quaternion.

## APPENDIX B. The mapping between $q_i$ and $\rho_i$, $i = \overline{1,3}$

The kinematic scheme of the first kinematic chain KC₁ is shown in Figure B1.a, whereas the relative motions between the mechanism components are shown in Figure B1.b (where motion arrows without label represent passive motions).

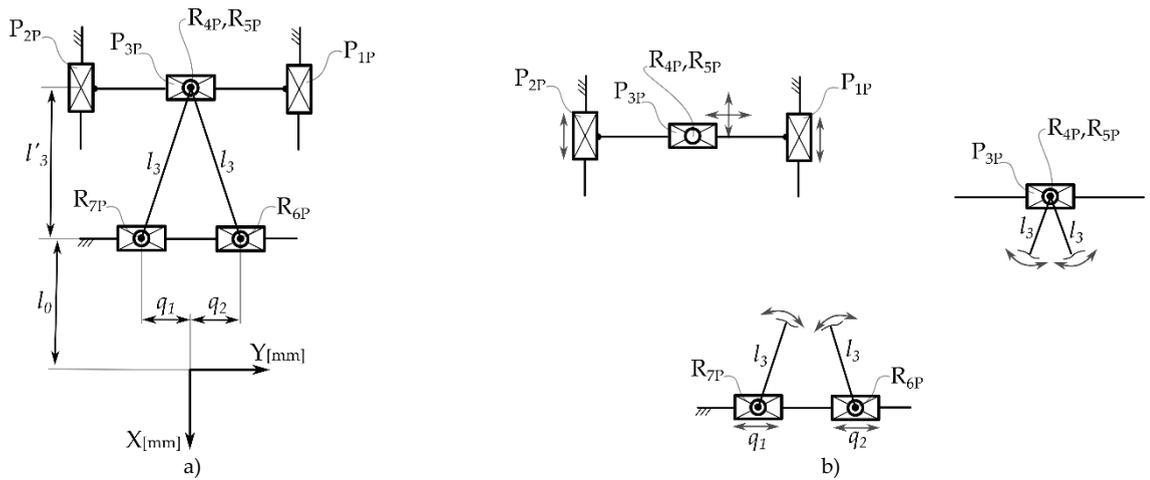

**Figure B1**. The kinematic scheme of KC₁: a) the planar mechanism; b) detail on the relative motion between components.

As shown in Figure B1.a, the height of the triangle of sides $l_3$ and $|q_1| + |q_2|$ is $l'_3$ which can be computed through:

$$l'_3 = \sqrt{l_3^2 - \left(\tfrac{1}{2}q_2 - \tfrac{1}{2}q_1\right)^2}. \tag{B1}$$

The equivalent kinematic scheme of the second kinematic chain KC₂ is shown in Figure B2.a, posing in the *Y0Z* plane; for KC₂ the relative position between mechanical elements is equivalent for any plane rotated around the *OY* axis, the *Y0Z* plane is selected for simplicity. The relative motions between the mechanism components are shown in Figure B2.b.

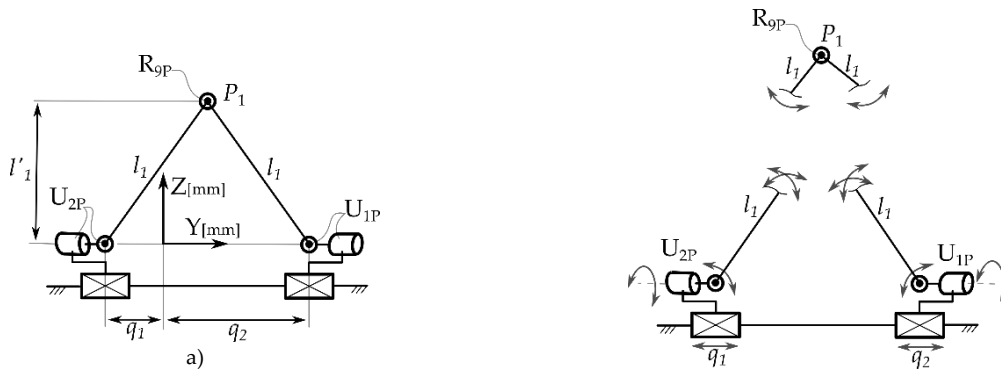

**Figure B2**. The kinematic scheme of KC2: a) the mechanism posed in the YOZ plane; b) detail on the relative motion between components.

The height of the triangle with sides $l_1$ and $|q_1|+|q_2|$ is $l_1'$ which can be computed as:

$$l_1' = \sqrt{l_1^2 - \left(\tfrac{1}{2}q_2 - \tfrac{1}{2}q_1\right)^2}. \tag{B2}$$

It is important to note that $l_1'$ always intersects $l_3'$ at $\tfrac{1}{2}(q_2 - q_1)$ which is defined as the value of parameter $\rho_1$ (Figure 3). Consequently:

$$\rho_1 = \tfrac{1}{2}(q_1 + q_2). \tag{B3}$$

Furthermore, the value of parameter $\rho_2$ (Figure 3) is defined as $\rho_2 = l_1' + l_4$, consequently:

$$(\rho_2 - l_4)^2 = \left(\tfrac{1}{2}q_2 - \tfrac{1}{2}q_1\right)^2 - l_1^2. \tag{B4}$$

Since $l_1'$ always intersects $l_3'$ at $\tfrac{1}{2}(q_2 - q_1)$ an XOZ perspective of PM can be represented as in Figure B3. This representation also highlights the action of third kinematic chain KC3 on point $P(X_P, Y_P, Z_P)$. $R_{q3}$ is defined as the characteristic point of the active joint $q_3$. Furthermore, this perspective highlights the definition of the $\rho_3$ parameter (Figure 3), which is an angle between the ZOY plane and the plane that contains KC2.

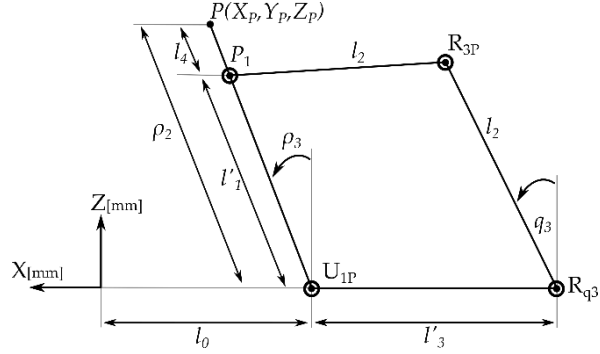

**Figure B3**. The kinematic scheme of KC3 and its effect on point $P(X_P, Y_P, Z_P)$

To derive the connection between $\rho_3$ and $q_3$ parameters two vectors can be constructed in a XOZ plane using the characteristic points in the revolute joints in Figure B3:

$$\mathbf{V}_1 \equiv \overrightarrow{U_{1P}P_1} = \begin{bmatrix} l_1' \sin(\rho_3) \\ l_1' \cos(\rho_3) \end{bmatrix}, \mathbf{V}_2 \equiv \overrightarrow{R_{q3}R_{3P}} = \begin{bmatrix} l_3' + l_2 \sin(q_3) \\ l_2 \cos(q_3) \end{bmatrix}, \tag{B5}$$

and the dot product $(\mathbf{V}_2 - \mathbf{V}_1) \cdot (\mathbf{V}_2 - \mathbf{V}_1) = l_2^2$ is used to define an input-output relation with $\rho_3$ and $q_3$ as unknowns. The final input-output relations for the $q_i, i = \overline{1,3}$ and $\rho_i, i = \overline{1,3}$ are obtained from Eqs. (B1) – (B5):

$$IO_{\rho\_q} : \begin{cases} \rho_1 - \tfrac{1}{2}(q_1 + q_2) = 0 \\ \left(\tfrac{1}{2}q_1 + \tfrac{1}{2}q_2\right)^2 + (\rho_2 - l_4)^2 - l_1^2 = 0 \\ \left(l_3' - l_2 \sin(q_3) + l_1' \sin(\rho_3)\right)^2 + \left(l_2 \cos(q_3) - l_1' \cos(\rho_3)\right)^2 - l_2^2 = 0 \end{cases}, \tag{B6}$$

$$l_1' = \sqrt{l_1^2 - \left(\tfrac{1}{2}q_1 + \tfrac{1}{2}q_2\right)^2}, \; l_3' = \sqrt{l_3^2 - \left(\tfrac{1}{2}q_1 + \tfrac{1}{2}q_2\right)^2},$$

**APPENDIX C. Numerical examples**

For the numerical examples the following geometric values are used: $[l = 400, l_0 = 300, l_1 = 200, l_2 = 150, l_3 = 170, l_4 = 50]$ where all dimensions are defined in *mm*.

*C.1 Numerical examples for the mapping between $q_i$ and $\rho_i$, $i = \overline{1,3}$*

The mapping between $q_i$ and $\rho_i$, $i = \overline{1,3}$ parameters is defined in Eq. (18). For the inverse kinematics the equation is solved for the $q_i$, $i = \overline{1,3}$ parameters and the $\rho_i$, $i = \overline{1,3}$ parameters are considered inputs; to illustrate an example the following values are defined: $\rho_1 = 50\,mm$, $\rho_2 = 180\,mm$, $\rho_3 = \frac{1}{3}\pi$. Table C1 shows the numerical solutions for the 4 inverse kinematic solutions (rounded at 3 decimal places).

**Table C1.** Inverse kinematic solutions for the mapping between $q_i$ and $\rho_i$, $i = \overline{1,3}$ parameters.

| Sol. No. | $q_1\,[mm]$ | $q_2\,[mm]$ | $q_3\,[rad]$ |
|---|---|---|---|
| 1 | −101.987 | 201.987 | 0.396 |
| 2 | −101.987 | 201.987 | 2.082 |
| 3 | 201.987 | −101.987 | 2.082 |
| 4 | 201.987 | −101.987 | 0.396 |

For the forward kinematics Eq. (18) is solved for the $\rho_i$, $i = \overline{1,3}$ parameters, therefore the $q_i$, $i = \overline{1,3}$ parameters are considered inputs; for a numeric example the values of Sol. No. 1 from Table C1 are chosen. The numerical values for the $\rho_i$, $i = \overline{1,3}$ parameters (up to 3 decimal places) are presented in Table C2.

**Table C2.** Forward kinematic solutions for the mapping between $q_i$ and $\rho_i$, $i = \overline{1,3}$ parameters.

| Sol. No. | $\rho_1\,[mm]$ | $\rho_2\,[mm]$ | $\rho_3\,[rad]$ |
|---|---|---|---|
| 1 | 50 | 179.999 | 1.047 |
| 2 | 50 | 179.999 | −1.309 |
| 3 | 50 | −79.999 | −1.309 |
| 4 | 50 | −79.999 | 1.047 |

The first solutions from Tables C1 and C2 represent the desired assembly mode and working mode for the modular parallel robot.

*C.2 Numerical examples for the mapping between $\rho_i$, $i = \overline{1,3}$ and $\mathbf{E}[X_E, Y_E, Z_E]^T$ (geometric approach)*

For the inverse kinematics Eqs. (23), (22) and (21) are solved for $[\psi, \theta, l_{ins}]$, $[X_P, Y_P, Z_P]$, and $\rho_i$, $i = \overline{1,3}$ parameters, respectively. Therefore, the inputs are $[X_E, Y_E, Z_E]$ (the Cartesian coordinates of the surgical instrument tip), and the outputs are back substituted until the values of $\rho_i$, $i = \overline{1,3}$ are computed. To illustrate a numeric example the following values are chosen $[X_E = 20, Y_E = 20, Z_E = -30]$ (all dimensions are defined in *mm*). Table C3.a shows the numerical values of the RCM parameters $[\psi, \theta, l_{ins}]$; the first solution in Table C3.a (the intended assembly mode) is subsequently substituted in the solutions of Eq. (22) for the $[X_P, Y_P, Z_P]$ parameters, and the numerical values are reported in Table C3.b; lastly the solution from Table C3.b are substituted into the solutions of Eq. (21) for the $\rho_i$, $i = \overline{1,3}$ parameter, and the results are reported in Table C3.c.

**Table C3.a** Inverse kinematic solutions for the mapping between $[X_E, Y_E, Z_E]$ and $[\psi, \theta, l_{ins}]$ parameters.

| Sol. No. | $\psi\,[rad]$ | $\theta\,[rad]$ | $l_{ins}\,[mm]$ |
|---|---|---|---|
| 1 | 0.785 | 0.81 | 41.231 |
| 2 | 0.785 | −2.327 | −41.231 |
| 3 | −2.356 | −0.81 | −41.231 |
| 4 | −2.356 | 2.327 | 41.231 |

**Table C3.b** Inverse kinematic solutions for the mapping between $[\psi, \theta, l_{ins}]$ and $[X_P, Y_P, Z_P]$ parameters.

| Sol. No. | $X_P$ [mm] | $Y_P$ [mm] | $X_P$ [mm] |
|---|---|---|---|
| 1 | −174.028 | −174.028 | 261.043 |

**Table C3.c** Inverse kinematic solutions for the mapping between $[X_P, Y_P, Z_P]$ and $\rho_i$, $i = \overline{1,3}$ parameters.

| Sol. No. | $\rho_1$ [mm] | $\rho_2$ [mm] | $\rho_3$ [rad] |
|---|---|---|---|
| 1 | −174.028 | 289.848 | 0.449 |
| 2 | −174.028 | −289.848 | −2.692 |

For the forward kinematics the first solution form Table C3.c is chosen and substituted into the solutions of Eq. (21) (solved for $[X_P, Y_P, Z_P]$). The numerical solutions are already reported in Table C3.b; furthermore, the numerical solutions for $[X_P, Y_P, Z_P]$ are substituted into the solutions of Eq. (22) (solved for $[\psi, \theta, l_{ins}]$), the numerical values being presented in Table C4.a. The first solution from Table C4.a is substituted in the solutions of Eq. (23) (solved for $[X_E, Y_E, Z_E]$) yielding unique solution which is presented in Tabled C4.b

**Table C4.a** Forward kinematic solutions for the mapping between $[X_P, Y_P, Z_P]$ and $[\psi, \theta, l_{ins}]$ parameters.

| Sol. No. | $\psi$ [rad] | $\theta$ [rad] | $l_{ins}$ [mm] |
|---|---|---|---|
| 1 | 0.785 | 0.81 | 41.231 |
| 2 | −2.356 | 2.327 | 41.231 |
| 3 | −2.356 | −0.81 | 758.767 |
| 4 | 0.785 | −2.327 | 758.767 |

**Table C4.b** Forward kinematic solutions for the mapping between $[X_P, Y_P, Z_P]$ and $[\psi, \theta, l_{ins}]$ parameters.

| Sol. No. | $X_E$ [mm] | $Y_E$ [mm] | $X_E$ [mm] |
|---|---|---|---|
| 1 | 20 | 20 | −30 |

*C.3 Numerical examples for the mapping between $\rho_i$, $i = \overline{1,3}$ and $\mathbf{E}[X_E, Y_E, Z_E]^T$ (homogeneous matrix approach)*

For the inverse kinematic model Eqs. (28) and (30) are considered with the vector relation $\mathbf{P}_2 - \mathbf{P}_2 = \mathbf{0}$, which yield input-output relations between $\rho_i$, $i = \overline{1,3}$ and $[\psi, \theta, l_{ins}]$ parameters. Solving $\mathbf{P}_2 - \mathbf{P}_2 = \mathbf{0}$ for $\rho_i$, $i = \overline{1,3}$ (for inverse kinematics) and using the first solution from Table C4.a yields numerical solution for $\rho_i$, $i = \overline{1,3}$ which are already presented in Table C3.c. Solving $\mathbf{P}_2 - \mathbf{P}_2 = \mathbf{0}$ for $[\psi, \theta, l_{ins}]$ (for forward kinematics) and using the first solution from Table C3.c yields numerical solution for $[\psi, \theta, l_{ins}]$ which are already presented in Table C4.a. At this point the $[\psi, \theta, l_{ins}]$ parameters are known, and to obtain the coordinates of the instrument tip $\mathbf{E}[X_E, Y_E, Z_E]^T$, Eq. (31) is used with inputs the numerical values of the first solution in Table C4.a.

*C.4 Numerical examples for the mapping between $\rho_i$, $i = \overline{1,3}$ and $\mathbf{Q}_E[x_{iE}, y_{iE}]^T$, $i = \overline{0,3}$ (dual quaternion approach)*

For the inverse kinematics Eq. (39) is used, which represents the solutions for the $\rho_1, \rho_2, u_3$ parameters ($u_3$ being the tangent of half angle of $\rho_3$) from the Groebner basis $G_{IK}$ discussed in Sub-Section 4.4. For a numerical example the following input parameters are used: $u_1 = 0.4142$, $u_2 = 0.4316$, $l_{ins} = 41.231 mm$, where $u_1, u_2$ are the tangents of half angles of $\psi, \theta$ parameters. The numerical results for evaluating Eq. (39) with the numerical values for the $u_1, u_2, l_{ins}$ parameters are shown in Table C5.a. For the forward kinematics the solution of the Groebner base $G_{FK}$ (also discussed in Sub-Section 4.4) solved for the $u_1, u_2, l_{ins}$ are evaluated with the first solution of Table C5.a, yielding the numerical values shown in Table C5.b. Note that the solutions for both inverse and forward kinematics are doubled for the kinematic models derived with the dual quaternion approach.

**Table C5.a** Inverse kinematic solutions for the mapping between $u_1, u_2, l_{ins}$ and $\rho_1, \rho_2, u_3$ parameters.

| Sol. No. | $\rho_1$ [mm] | $\rho_2$ [mm] | $u_3$ |
|---|---|---|---|
| 1,2 | −174.028 | 289.848 | 0.449 |
| 3,4 | −174.028 | −289.848 | −4.373 |

**Table C5.b** Forward kinematic solutions for the mapping between $u_1, u_2, l_{ins}$ and $\rho_1, \rho_2, u_3$ parameters.

| Sol. No. | $u_1$ | $u_2$ | $l_{ins}$ [mm] |
|---|---|---|---|
| 1,2 | 0.4142 | 0.4316 | 41.231 |

| 3,4 | −2.4142 | 2.317 | 41.231 |
| 5,6 | −2.4142 | −0.4316 | 758.766 |
| 7,8 | 0.4142 | −2.317 | 758.766 |

# References

[1] Topal H, Aerts R, Laenen A, et al. Survival After Minimally Invasive vs Open Surgery for Pancreatic Adenocarcinoma. JAMA Netw Open. 5(12) (2022) e2248147. doi:10.1001/jamanetworkopen.2022.48147

[2] J.D. Mizrahi, R. Surana, J.W. Valle, R.T. Shroff, Pancreatic cancer Lancet, 395 (10242) (2020) 2008-2020. 10.1016/S0140-6736(20)30974-0

[3] B. A. Uijterwijk, et al. International Study Group on non-pancreatic periampullary Cancer (ISGACA) (2024). Oncological resection and perioperative outcomes of robotic, laparoscopic and open pancreatoduodenectomy for ampullary adenocarcinoma: a propensity score matched international multicenter cohort study. HPB: the official journal of the International Hepato Pancreato Biliary Association, S1365-182X(24) (2020) 02428-6. https://doi.org/10.1016/j.hpb.2024.11.013

[4] C. Vaida, N. Plitea, D. Pisla, B. Gherman, Orientation module for surgical instruments—a systematical approach. Meccanica 48 (2013) 145–158. https://doi.org/10.1007/s11012-012-9590-x

[5] W. Zhang, Z. Wang, K. Ma, et al., State of the art in movement around a remote point: a review of remote center of motion in robotics. Front. Mech. Eng. 19 (14) (2024). https://doi.org/10.1007/s11465-024-0785-3

[6] L. Rongfu, G. Weizhong, C. Shing Shin, Type synthesis of 2R1T remote center of motion parallel mechanisms with a passive limb for minimally invasive surgical robot, Mech. Mach. Theory, 172, 104766, June 2022. https://doi.org/10.1016/j.mechmachtheory.2022.104766

[7] M. Zhou, Q. Yu, K. Huang, S. Mahov, A. Eslami, M. Maier, C.P. Lohmann, N. Navab, D. Zapp, A. Knoll, M.A. Nasseri, Towards robotic-assisted subretinal injection: a hybrid parallel–serial robot system design and preliminary evaluation, ITIE, 67 (2020) 6617-6628. doi: 10.1109/TIE.2019.2937041

[8] G. Chen, J. Wang, H. Wang, C. Chen, V. Parenti-Castelli, J. Angeles, Design and validation of a spatial two-limb 3R1T parallel manipulator with remote center-of-motion, Mech. Mach. Theory, 149 (2020), Article 103807. https://doi.org/10.1016/j.mechmachtheory.2020.103807

[9] G. Chen, J. Wang, H. Wang, A new type of planar two degree-of-freedom remote center-of-motion mechanism inspired by the Peaucellier–Lipkin straight-line linkage, J. Mech. Des., 141 (1) (2018), 015001. https://doi.org/10.1115/1.4041221

[10] A. Yaşır, G. Kiper, M.İ.C. Dede, Kinematic design of a non-parasitic 2R1T parallel mechanism with remote center of motion to be used in minimally invasive surgery applications, Mech. Mach. Theory, 153 (2020), Article 104013. https://doi.org/10.1016/j.mechmachtheory.2020.104013

[11] H. Long, Y. Yang, X. Jingjing, S. Peng, Type Synthesis of 1R1T Remote Center of Motion Mechanisms Based on Pantograph Mechanisms, J. Mech. Des., 138 (2016), Article 01450. https://doi.org/10.1115/1.4031804

[12] D. Chablat, G. Michel, P. Bordure, S. Venkateswaran, R. Jha, Workspace analysis in the design parameter space of a 2-DOF spherical parallel mechanism for a prescribed workspace: Application to the otologic surgery. Mech. Mach. Theory, 157 (2021) 104224. https://doi.org/10.1016/j.mechmachtheory.2020.104224

[13] L. Li, D. Zhang, C. Tian, A family of generalized single-loop RCM parallel mechanisms: structure synthesis, kinematic model, and case study, Mech. Mach. Theory., 195 (2024), Article 105606. https://doi.org/10.1016/j.mechmachtheory.2024.105606

[14] Calin Vaida, Bogdan Gherman, Paul Tucan, Iosif Birlescu, Damien Chablat, Doina Pisla,. Parallel robotic system for minimally invasive surgery of the pancreas, Patent pending A00522/11.09.2024.

[15] S. Xu, M. Chu, H. Sun, Design and Stiffness Optimization of Bionic Docking Mechanism for Space Target Acquisition. Appl. Sci. 11 (2021) 10278. 10.3390/app112110278

[16] R. Santoro, M. Pontani, Orbit acquisition, rendezvous, and docking with a noncooperative capsule in a Mars sample return mission, Acta Astronautica, 211 (2023) 950-962. https://doi.org/10.1016/j.actaastro.2023.04.043

[17] T. Sun, C. Liu, B. Lian, P. Wang and Y. Song, Calibration for Precision Kinematic Control of an Articulated Serial Robot, in IEEE Transactions on Industrial Electronics, 68(7) (2021) 6000-6009. doi: 10.1109/TIE.2020.2994890

[18] L. Rapetti, Y. Tirupachuri, K. Darvish, S. Dafarra, G. Nava, C. Latella, D. Pucci, Model-Based Real-Time Motion Tracking Using Dynamical Inverse Kinematics. Algorithms, 13 (2020) 266. 10.3390/a13100266

[19] L. Berscheid, T. Kroeger, Jerk-limited Real-time Trajectory Generation with Arbitrary Target States, Robotics: Science and Systems 2021. DOI: 10.15607/RSS.2021.XVII.015

[20] E. Ferrentino, H. J. Savino, A. Franchi and P. Chiacchio, A Dynamic Programming Framework for Optimal Planning of Redundant Robots Along Prescribed Paths With Kineto-Dynamic Constraints, in IEEE Transactions on Automation Science and Engineering, 21(4) (2023). doi:10.1109/TASE.2023.3330371

[21] P. Wu, T. He, H. Zhu, Y. Wang, Q. Li, Z. Wang, X. Fu, F. Wang, P. Wang, C. Shan, Z. Fan, L. Liao, P. Zhou, W. Hu, Next-generation machine vision systems incorporating two-dimensional materials: Progress and perspectives, InfoMat, 4(1) (2021). https://doi.org/10.1002/inf2.12275.

[22] A. Nazari, K. Zareinia, F. Janabi-Sharifi, Visual servoing of continuum robots: Methods, challenges, and prospects, The International Journal of Medical Robotics and Computer Assisted Surgery, 18(3) (2022). https://doi.org/10.1002/rcs.2384

[23] J. Li, Q. Xue, S. Yang, X. Han, S. Zhang, M. Li, J. Guo. Kinematic analysis of the human body during sit-to-stand in healthy young adults. Medicine (Baltimore), 100(22) (2021) e26208. doi:10.1097/MD.0000000000026208

[24] B.T. van Oeveren, C.J. de Ruiter, P.J. Beek, J.H. van Dieën. The biomechanics of running and running styles: a synthesis. Sports Biomechanics, 23(4) (2021) 516–554. https://doi.org/10.1080/14763141.2021.1873411

[25] C. Gosselin, J. Angeles, Singularity analysis of closed-loop kinematic chains, IEEE Trans. Robot. Autom. 6 (1990) 281–290. doi: 10.1109/70.56660

[26] J. Angeles, D. Chablat, On Isotropic Sets of Points in the Plane. Application to the Design of Robot Architectures. In: Lenarčič, J., Stanišić, M.M. (eds) Advances in Robot Kinematics. Springer, Dordrecht. (2000) 73-82. https://doi.org/10.1007/978-94-011-4120-8_8

[27] L. Szűcs, P. Galambos and D. A. Drexler, Kinematics of Delta-type Parallel Robot Mechanisms via Screw Theory: A tutorial paper, 2021 IEEE 19th World Symposium on Applied Machine Intelligence and Informatics (SAMI), Herl'any, Slovakia, pp. 000293-000298, 2021.

[28] A. Müller, An overview of formulae for the higher-order kinematics of lower-pair chains with applications in robotics and mechanism theory, Mechanism and Machine Theory, 142, (2019) 103594. https://doi.org/10.1016/j.mechmachtheory.2019.103594

[29] M. Pfurner, HP Schröcker, M. Husty, Path Planning in Kinematic Image Space Without the Study Condition. In: Lenarčič, J., Merlet, JP. (eds) Advances in Robot Kinematics 2016. Springer Proceedings in Advanced Robotics, Springer, Cham, 4 (2017) 285-292. https://doi.org/10.1007/978-3-319-56802-7_30

[30] Z. Fu, J. Pan, E. Spyrakos-Papastavridis, X. Chen, M. Li, A Dual Quaternion-Based Approach for Coordinate Calibration of Dual Robots in Collaborative Motion, in IEEE Robotics and Automation Letters, 5(3) (2020) 4086-4093. doi:10.1109/LRA.2020.2988407

[31] N.T. Dantam, Robust and efficient forward, differential, and inverse kinematics using dual quaternions. The International Journal of Robotics Research, 40(10-11) (2020) 1087-1105. https://doi.org/10.1177/0278364920931948

[32] D. Condurache, Higher-Order Relative Kinematics of Rigid Body and Multibody Systems, A Novel Approach with Real and Dual Lie Algebras, Mech. Mach. Theory 176 (2022) 104999. https://doi.org/10.1016/j.mechmachtheory.2022.104999

[33] J. Funda, R. Paul, A computational analysis of screw transformations in robotics, IEEE Trans. Robot. Autom. 3 (1990) 348–356. doi: 10.1109/70.56653

[34] D. Condurache, Higher-Order Kinematics of Lower-Pair Chains With Hyper-Multidual Algebra, Proceedings of the ASME 2022 International Design Engineering Technical Conferences and Computers and Information in Engineering Conference. Volume 7: 46th Mechanisms and Robotics Conference (MR). St. Louis, Missouri, USA. August 14–17, 2022. V007T07A073. ASME, November 2022.

[35] D. Condurache, A. Burlacu, Dual tensors based solutions for rigid body motion parameterization, Mechanism and Machine Theory, 74 (2014) 390-412. https://doi.org/10.1016/j.mechmachtheory.2013.12.016

[36] D. Condurache, M. Cojocari, I. Birlescu, B. Gherman, Automatic Differentiation of Serial Manipulator Jacobians Using Multidual Algebra, Mechanisms and Machine Science, 157. Springer, Cham, 2024. https://doi.org/10.1007/978-3-031-59257-7_38

[37] M. Husty, I. Birlescu, P. Tucan, C. Vaida, D. Pisla, An algebraic parameterization approach for parallel robots analysis, Mech. Mach. Theory, 140 (2019) 245-257. https://doi.org/10.1016/j.mechmachtheory.2019.05.024

[38] N. Plitea, J. Hesselbach, D. Pisla, A. Raatz, C. Vaida, J. Wrege, A. Burisch, Innovative development of parallel robots and microrobots, Acta Tech. Napoc. Ser. Appl. Math. Mec. 49, 5–26 2006

[39] D.A. Cox, J. Little, D. O'Shea, Ideals, Varieties, and Algorithms, An Introduction to Computational Algebraic Geometry and Commutative Algebra. Springer Cham, 2015.

[40] A. Müller, S. Kumar, T. Kordik, A recursive lie-group formulation for the second-order time derivatives of the inverse dynamics of parallel kinematic manipulators. IEEE Robot. Automat. Lett. 8 (2023) 3804–3811. doi: 10.1109/LRA.2023.3267005